\documentclass[10pt]{article}  
\usepackage{times}
\newcommand{\circnum}[1]
{\setlength{\unitlength}{0.00083300in}%
\begin{picture}(210,210)(0,40)
\put(105,105){\circle{210}}
\put(105,105){\makebox(0,0){#1}}
\end{picture}
}
\usepackage{forest}
\usepackage{amsmath}
\usepackage{amssymb}  
\usepackage{latexsym}  
\usepackage{array}   

\makeatletter
\DeclareRobustCommand\bigop[1]{%
  \mathop{\vphantom{\sum}\mathpalette\bigop@{#1}}\slimits@
}
\newcommand{\bigop@}[2]{%
  \vcenter{%
    \sbox\z@{$#1\sum$}%
    \hbox{\resizebox{\ifx#1\displaystyle.9\fi\dimexpr\ht\z@+\dp\z@}{!}{$\m@th#2$}}%
  }%
}
\makeatother

\begingroup\makeatletter
\gdef\mydoublespacing{\ifcase \@ptsize \relax 
    \def\baselinestretch{1.667}
  \or 
    \def\baselinestretch{1.618}
  \or 
    \def\baselinestretch{1.655}
  \fi
 \ifx\undefined\@newbaseline
  \ifx\@currsize\normalsize\@normalsize\else\@currsize\fi%
 \else
  \@newbaseline%
\fi%
}
\endgroup

\begingroup\makeatletter
\gdef\mysinglespacing{\def\baselinestretch{1}
 \ifx\undefined\@newbaseline
  \ifx\@currsize\normalsize\@normalsize\else\@currsize\fi%
 \else
  \@newbaseline%
\fi%
  \vskip\baselineskip
}
\endgroup
\def\belowmat#1#2{\raise-15pt\hbox{$\smash{\begin{array}{c}#1 \\[10pt]
        {\scriptsize \mbox{$#2$}} \end{array}}$}}

\def\Rbb{\mathbb{R}}

\newcommand{\trace}{\operatorname{tr}}

\newcommand{\argmin}{\text{arg}\min}

\newcommand{\rank}{\operatorname{rank}}

\def\<{\hbox{$\langle$}}
\def\>{\hbox{$\rangle$}}

\newcommand{\matsize}[2]{\ensuremath{#1\times #2}}

\def\cbf{\mathbf{c}} 
 
\def\ebf{\mathbf{e}}

\def\vbf{\mathbf{v}}

\def\cbftilde{\tilde{\cbf}}

\def\Vtilde{\tilde{V}}

\def\zerobf{\mbox{\bf 0}}

\def\ihat{\hat{i}}

\newsavebox{\factboxbox}                    
{\begin{lrbox}{\factboxbox}
}
{\end{lrbox}\begin{center}\fbox{\usebox{\factboxbox}}\end{center}
}

\makeatletter
{\newdimen\p@renwd
\p@renwd=20pt
\gdef\bbordermatrix#1{\begingroup \m@th
  \setbox\z@\vbox{\def\cr{\crcr\noalign{\kern2\p@\global\let\cr\endline}}%
    \ialign{$##$\hfil\kern2\p@\kern\p@renwd&\thinspace\hfil$##$\hfil
      &&\quad\hfil$##$\hfil\crcr
      \omit\strut\hfil\crcr\noalign{\kern-\baselineskip}%
      #1\crcr\omit\strut\cr}}%
  \setbox\tw@\vbox{\unvcopy\z@\global\setbox\@ne\lastbox}%
  \setbox\tw@\hbox{\unhbox\@ne\unskip\global\setbox\@ne\lastbox}%
  \setbox\tw@\hbox{$\kern\wd\@ne\kern-\p@renwd\left[\kern-\wd\@ne
    \global\setbox\@ne\vbox{\box\@ne\kern2\p@}%
    \vcenter{\kern-\ht\@ne\unvbox\z@\kern-\baselineskip}\,\right]$}%
  \null\;\vbox{\kern\ht\@ne\box\tw@}\endgroup}
}
\makeatother

\makeatother%

\newsavebox{\citenametkm}                    
\newsavebox{\citesource}
\def\quotesourcefont{\small}

\newlength{\exmatsp} 
\newlength{\Exmatsp}
\def\syncsep{\kern -.3ex \raisebox{2ex}{\makebox[0cm]{$\downarrow$}}\kern .3ex}

\newcounter{stepctr}

\makeatletter
\newif\if@borderstar
\def\bordermatrix{\@ifnextchar*{%
\@borderstartrue\@bordermatrix@i}{\@borderstarfalse\@bordermatrix@i*}%
}
\def\@bordermatrix@i*{\@ifnextchar[{\@bordermatrix@ii}{\@bordermatrix@ii[()]}}
\def\@bordermatrix@ii[#1]#2{%
\begingroup
\m@th\@tempdima8.75\p@\setbox\z@\vbox{%
\def\cr{\crcr\noalign{\kern 2\p@\global\let\cr\endline }}%
\ialign {$##$\hfil\kern 2\p@\kern\@tempdima & \thinspace %
\hfil $##$\hfil && \quad\hfil $##$\hfil\crcr\omit\strut %
\hfil\crcr\noalign{\kern -\baselineskip}#2\crcr\omit %
\strut\cr}}%
\setbox\tw@\vbox{\unvcopy\z@\global\setbox\@ne\lastbox}%
\setbox\tw@\hbox{\unhbox\@ne\unskip\global\setbox\@ne\lastbox}%
\setbox\tw@\hbox{%
$\kern\wd\@ne\kern -\@tempdima\left\@firstoftwo#1%
\if@borderstar\kern2pt\else\kern -\wd\@ne\fi%
\global\setbox\@ne\vbox{\box\@ne\if@borderstar\else\kern 2\p@\fi}%
\vcenter{\if@borderstar\else\kern -\ht\@ne\fi%
\unvbox\z@\kern-\if@borderstar2\fi\baselineskip}%
\if@borderstar\kern-2\@tempdima\kern2\p@\else\,\fi\right\@secondoftwo#1 $%
}\null \;\vbox{\kern\ht\@ne\box\tw@}%
\endgroup
}
\makeatother

\usepackage{subcaption}
\usepackage{alltt}
\usepackage{multicol}
\usepackage{verbatim}
\usepackage{url}
\usepackage{floatrow}
\newfloat{algorithm}{thp}{loa}  
\floatname{algorithm}{Algorithm}  

\def\q{\hspace*{10pt}}

\begin{document}

\title{Document Author Classification Using Parsed Language Structure}

\author{Todd K. Moon, Jacob H. Gunther \\
  Electrical and Computer Engineering Department, Utah State
  University, Logan, Utah \\
  \texttt{ \{todd.moon, jake.gunther\} @ usu.edu }
}
\date{\null}
\maketitle

\begin{abstract}
  Over the years there has been ongoing interest in detecting
  authorship of a text based on statistical properties of the text,
  such as by using occurrence rates of noncontextual words.  In
  previous work, these techniques have been used, for example, to
  determine authorship of all of \emph{The Federalist Papers}.  Such
  methods may be useful in more modern times to detect fake or AI
  authorship.  Progress in statistical natural language parsers
  introduces the possibility of using grammatical structure to detect
  authorship.  In this paper we explore a new possibility for
  detecting authorship using grammatical structural information
  extracted using a statistical natural language parser.  This paper
  provides a proof of concept, testing author classification based on
  grammatical structure on a set of ``proof texts,'' \emph{The
    Federalist Papers} and \emph{Sanditon} which have been as test
  cases in previous authorship detection studies.  Several features
  extracted from the statistical natural language parser were
  explored: all subtrees of some depth from any level; rooted subtrees
  of some depth, part of speech, and part of speech by level in the
  parse tree.  It was found to be helpful to project the features into
  a lower dimensional space.  Statistical experiments on these
  documents demonstrate that information from a statistical parser
  can, in fact, assist in distinguishing authors.
\end{abstract}


\section{Introduction and Background}

There has been considerable effort over the years related to using
statistical methods to identify authorship of texts, based on examples
from candidate authors, in what is sometimes called ``stylometry'' or
``author identification.''  Statistical analysis of documents goes
back to Augustus de Morgan in 1851 \cite[p.{} 282]{Lord1958},
\cite[p.{} 166]{Morton1978}, who proposed that word length statistics
might be used to determine the authorship of the Pauline epistles.
Stylometry was employed as early as 1901 to explore the authorship of
Shakespeare \cite{Mendenhall1901}.  Since then, it has been employed
in a variety of literary studies (see, e.g.
\cite{Chretien1964,WishartLeach1972,Brinegar1963}), including twelve
of {\em The Federalist Papers} which were of uncertain authorship
\cite{MostellerWallace1964} --- which we re-examine here --- and an
unfinished novel by Jane Austen ---which we also re-examine here.
Information theoretic techniques have also been used more recently
\cite{JanusJagenauer2005}.  Earlier work in stylometry has been based
on ``noncontextual words,'' words which do not convey the primary
meaning of the text, but which act in the background of the text to
provide structure and flow.  Noncontextual words are at least
plausible, since an author may address a variety of topics, so
particular distinguishing words are not necessarily revealing of
authorship.  In noncontextual word studies, a set of most common words
noncontextual is selected \cite{Morton1978}, and documents are
represented by word counts, or ratios of word counts to document
length.  A review of the statistical methods is in \cite{Holmes1985}.
As a variation, sets of ratios of counts of noncontextual word
patterns to other word patterns are also employed \cite{Hilton1990}.
Statistical analysis based on author vocabulary size {\em vs.}
document length --- the ``vocabulary richness'' --- has also been
explored \cite{Holmes1992}.  For other related work, see
\cite{LuyckxDaelemans2005,Grieve2007,IqbalBinsalleehFungDebbabi2007,Stamatos2009}

A more recent paper \cite{ZhangWuNiuDing2014} considers the
effectiveness of a wide variety of feature sets.  Feature sets
considered there include: vectors comprising frequencies of pronouns;
function words (that is, articles, pronouns, particles, expletives);
part of speech (POS); most common words; syntactic features (such as
noun phrase, or verb phrase); or tense (e.g. use of present or past
tense); voice (active of passive).  In \cite{ZhangWuNiuDing2014},
feature vectors are formed from combinations of histograms, then
reduced in dimensionality using a two-stage process of principle
component analysis \cite{Jolliffe1986} followed by dimension reduction
using linear discriminant analysis (LDA).  In their LDA, the
within-cluster scatter matrix is singular (due to the high dimension
of the feature vectors relative the number of available training
vectors), so their scatter matrix is regularized.  To test this, the
authors consider a range of regularization parameters, selecting one
which gives the best performance.

More recent work \cite{CorbaraChulviRossoMoreo2022} mentions the
survey in \cite{Stamatos2009} in which commonly used features in the
authorship field are word and character $n$-grams.  As noted, there
are risks the statistical methods might be biased by topic-related
patterns.  As \cite{CorbaraChulviRossoMoreo2022} observe, ``an
authorship classifier (even a seemingly good one) might end up
unintentionally performing topic identification if domain-dependent
features are used. ... In order to avoid this, researchers might limit
their scope to features that are clearly topic-agnostic, such as
function words or syntactic features.''  The work presented here falls
in the latter category, making use of grammatical structures
statistically extracted from the text.  These appear to be difficult
to spoof.  Examination of other recent works
\cite{CorbaraFerriolsRossoMoreo2022,SanchezPerezMarkovGomezAdorno2017}
indicate that there is ongoing interest in author identification
methods, but none making use of the grammatical structures use here;
there is a tendency to rely more on traditional $n$-grams.

In this work the feature vectors are obtained using tree information
from parse trees from a natural language parsing tool
\cite{KleinManning2003}.  These features were not among the features
considered in \cite{ZhangWuNiuDing2014}.  The grammatical structures
are, it seems, more subtle than simple counts of classes of words, and
hence may be less subject to spoofing or topic bias, since it seems
unlikely that an author intending to imitate another would be able to
coherently track complicated patterns of usage, and the features do
not include any words from the documents.  It is found that the
tree-based features perform better than the POS features on the test
data considered.

The feature vectors so obtained can be of very high dimension, so
dimension reduction is also performed here.  However, to deal with the
singularity of the within-cluster scatter matrix, a generalized SVD
approach is used, which avoids the need to select a regularization
parameter.


This paper provides a proof-of-concept of these tree-based features to
distinguish authorship by applying them to documents which have been
previously examined, \emph{The Federalist Papers} and \emph{Sanditon}.
The ability to classify by authorship is explored for several feature
vectors obtained from the parsed information.

\section{Statistical Parsing and Extracted Features}

Part-of-speech (POS) tagging classifies the words in a sentence
according to their part of speech, such as noun,verb, or interjection.
Because of the complexity of English language, there is potential for
ambiguity.  For example, many words (such as ``seat'' or ``bat'' or
``eye'') can be either nouns or verbs.  The ambiguity can be dealt
with using statistical parsing, in which a large corpus of language is
used to develop probabilistic models for words which are based on
contextual words.  These models are typically trained with the
assistance of human linguistic experts.  The parser used in this work
uses a language model developed using the annotated corpus called the
Penn Treebank, which is a corpus of over 7 million words of American
English, collected from multiple sources, labeled using human and
semi-automated markup
\cite{MarcusSantoriniMarcinkiewicz1993,Taylor03thepenn}.  The parser
is described in \cite{KleinManning2003}.  It is a probabilistic
context free grammar (PCFG) parser \cite{KleinManning2003PCFG}, with
language transition probabilities determined based on the Penn
Treebank corpus.  The parser software is known as the Stanford Parser
\cite{ParserPage}.  Parsing results presented here are produced by
version 4.2.0, released November 17, 2020.

Table \ref{tab:pos} lists the POS labels (the POS tagset) associated
with words when a sentence is parsed by this parser.  It also lists
the syntactic tagset, produced by the parser when doing grammatical
parsing.  (see \cite[Table 1.1, Table 1.2]{Taylor03thepenn},
\cite[Chapter 5]{JurafskyMartin2009}).

A brief introduction to statistical parsing is provide in Appendix A.

\begin{table}
  \captionsetup[table]{position=top}
\caption{Penn Treebank POS Tagset and Syntactic Tagset.}
  \label{tab:pos}
  \centering
  \resizebox{0.8\textwidth}{!}
{  
  \begin{tabular}{|lp{2in}|lp{2in}|}
\hline
\multicolumn{4}{|l|}{\textbf{POS tagset}  \cite[Table 1.2]{Taylor03thepenn}} \\
\hline
CC  & coordinating conjunction (and,but, or)  &
CD & cardinal number (one, two, three)  \\
DT & determiner (a, the) & EX & existential ``there'' \\
FW & foreign word & IN & preposition or subordinating conjunction (of,
                         in, by) \\
INTJ & interjection & JJ &  adjective (yellow) \\
JJR & adjective, comparative (bigger) & JJS & adjective, superlative
                                              (biggest) \\
LS & list item marker (1, 2, one) & 
MD & modal (can, should) \\
NN & noun, singular or mass (llama, snow) &
NNS & noun, plural (llamas) \\
NNP & proper noun, singular (IBM) &
NNPS & proper noun, plural (Carolinas) \\
PDT & predeterminer (all, both) &
POS & possessive ending ('s) \\
PRP & personal pronoun (I, you, he) &
PRP\$ & possessive pronoun (your, one's) \\
RB & adverb (quickly, never) &
RBR & adverb, comparative (faster) \\
RBS & adverb, superlative (fastest) &
RP & particle (up, off) \\
SYM & symbols ($+$, \%, \&) &
TO & ``to'' \\
UH & interjection (ah, oops) &
VB & verb, base form (eat) \\
VBD & verb, preterite (past tense) (ate) &
VBG & verb, gerund (eating) \\
VBN & verb, past participle (eaten) &
VBP & verb, non-3sg pre (eat) \\
VBZ & verb, 3sg pres (eats) &
WDT & wh-determiner (which, that) \\
WP & wh-pronoun (what, who) &
WP\$ & possessive WH- (whose) \\
WRB & wh-adverb (how, where) & &\\
\$ & dollar sign & \# & pound sign    \\
`` & left quote     & '' & right quote \\
( & left parenthesis & ) & right parenthesis \\
, & comma & . & sentence-final (. ! ?) \\
 : & mid-sentence punc (: : \ldots\  -- -)   & ${'}{'}$,~`,~' & straight double quote; left single open quote, right single close quote \\
\hline
    \multicolumn{4}{|l|}{\textbf{Syntactic Tagset} \cite[Table 1.1]{Taylor03thepenn}}\\
    \hline
ADJP & Adjective phrase & ADVP & Adverb phrase \\
NP     & Noun phrase      & PP      & Prepositional phrase \\
S       & Simple declarative clause & SBAR & Subordinate clause \\    
SBARQ & Direct question introduced by \emph{wh}-element & SINV  &
                                                                  Declarative
                                                                  sentence
                                                                  with
                                                                  subject-aux
                                                                  inversion \\
SQ   & Yes/no questions and subconstituent SBARQ excluding
       \emph{wh-element} & VP & Verb phrase \\
WHADVP & Wh-adverb phrase & WHNP & Wh-noun phrase \\
X & Constituent of unknown or uncertain category & * & ``Understood''
                                                       subject of
                                                       infinitive or
                                                       imperative \\    
0 & Zero variant of \emph{that} in subordinate clauses & T & Trace of
                                                            wh-Constituent
    \\
\hline    
  \end{tabular}
}
\end{table}

As an example of the parsing, consider the first sentence of
\emph{The Federalist Papers} 1 by Alexander Hamilton:
\begin{equation}
\begin{minipage}{.8\linewidth}
After an unequivocal experience of the inefficacy of the subsisting
 federal government, you are called upon to deliberate on a new
 Constitution for the United States of America.
\end{minipage}
\label{eq:text1}
\end{equation}
Parsing this sentences yields the tree representation portrayed in
figure \ref{fig:Tree_ex1}(a).  The leaf nodes correspond to the words
of the sentence, each labeled with a POS.  The non-leaf (interior)
nodes represent syntactic (grammatical structure) information
determined by the parser.  The label of each node of the tree is
referred to as a \emph{token}.  The parse tree can be represented
using the text string shown in \ref{fig:Tree_ex1}(b).  This is
formatted to show the various levels of the tree implied by the
nesting of the parentheses in figure \ref{fig:Tree_ex1}(c).

In preparing to extract feature vectors from a parse tree, some
additional tidying-up is performed.
The parser creates a ROOT node for each tree, which is therefore
uninformative and is removed.  Punctuation nodes in the tree, such as
(,\ ,), (.\ .), or (.\ ?) are removed.  Since the intent is to explore
how the parsed information can be used for classification, rather than
the words of the document, the words of the sentence are removed from
the parse tree.  With these edits, the sentence (\ref{eq:text1}) has
the parsed representation
\begin{equation}
  \label{eq:treeparse1}
\begin{minipage}{0.8\linewidth}
(S(PP(IN)(NP(NP(DT)(JJ)(NN))(PP(IN)(NP(NP(DT)

(NN))(PP(IN)(NP(DT)(JJ)(JJ)(NN)))))))

(NP(PRP))(VP(VBP)(VP(VBN)(PP(IN))

(S(VP(TO)(VP(VB)(PP(IN)(NP(NP(DT)(JJ)(NN))

(PP(IN)(NP(NP(DT)(NNP)(NNP))(PP(IN)(NP(NNP)))))))))))))
\end{minipage}
\end{equation}
From this prepared data, various feature vectors were extracted, as
described below.  (The text manipulation and data extraction was done
using the Python language, making extensive use of Python's dictionary
type.  The parsed string (\ref{eq:treeparse1}) can be used, for
example, as a key to a Python dictionary.)

\begin{figure}
\begin{center}
\begin{subfigure}{\textwidth}
\raisebox{-.2in}{
\resizebox{0.8\textwidth}{!} {
\begin{forest}[ROOT[S[PP[IN After][NP[NP[DT an][JJ unequivocal][NN
   experience]][PP[IN of][NP[NP[DT the][NN inefficacy]][PP[IN of][NP[DT
   the][JJ subsisting][JJ federal][NN government]]]]]]][,,][NP[PRP
   you]][VP[VBP are][VP[VBN called][PP[IN upon]][S[VP[TO to][VP[VB
   deliberate][PP[IN on][NP[NP[DT a][JJ new][NN Constitution]][PP[IN
   for][NP[NP[DT the][NNP United][NNP States]][PP[IN of][NP[NNP America]]]]]]]]]]]]]]
 \end{forest}
}
}
\caption{Graphical representation of parse tree}
\end{subfigure}
\end{center}

\begin{subfigure}{0.25\textwidth}
\raisebox{-.8in}{
\begin{minipage}{\textwidth}
\raggedright (ROOT(S
(PP(IN After)(NP(NP(DT an)(JJ unequivocal)(NN experience))(PP(IN of)(NP(NP(DT the)(NN inefficacy))
(PP(IN of)(NP(DT the)(JJ subsisting)(JJ federal)(NN government)))))))(, ,)(NP(PRP you))(VP(VBP are)
(VB(VBN called)(PP(IN upon))(S(VP(TO to) (VP(VB deliberate) (PP (IN on) (NP (NP(DT a)(JJ new)(NN Constitution))
(PP(IN for)(NP(NP(DT the)(NNP United)(NNP States))(PP(IN of)(NP(NNP
America))))))))))))))
\end{minipage}
}
\caption{Textual representation of parse tree}
\end{subfigure}
\qquad 
\begin{subfigure}{.7\textwidth}
{\tiny
\begin{minipage}[t]{\textwidth}
(ROOT\\
\q(S\\
\q\q(PP\\
\q\q\q(IN After)\\
\q\q\q(NP\\
\q\q\q\q(NP\\
\q\q\q\q\q(DT an)(JJ unequivocal)(NN experience))\\
\q\q\q\q(PP\\
\q\q\q\q\q(IN of)\\
\q\q\q\q\q(NP\\
\q\q\q\q\q\q(NP\\
\q\q\q\q\q\q\q(DT the)(NN inefficacy))\\
\q\q\q\q\q\q(PP\\
\q\q\q\q\q\q\q(IN of)\\
\q\q\q\q\q\q\q(NP\\
\q\q\q\q\q\q\q\q(DT the)(JJ subsisting)(JJ federal) \\
\q\q\q\q\q\q\q\q(NN government)))))))\\
\q\q(, ,)\\
\q\q(NP\\
\q\q\q(PRP you))\\
\q\q(VP\\
\q\q\q(VBP are)\\
\q\q(VB\\
\q\q\q\q(VBN called)\\
\q\q\q\q(PP\\
\q\q\q\q\q(IN upon))\\
\q\q\q\q(S\\
\q\q\q\q\q(VP\\
\q\q\q\q\q\q(TO to)\\
\q\q\q\q\q (VP\\
\q\q\q\q\q\q\q(VB deliberate)\\
\q\q\q\q\q (PP\\
\q\q\q\q\q\q\q (IN on)\\
\q\q\q\q\q (NP\\
\q\q\q\q\q\q\q (NP\\
\q\q\q\q\q\q\q\q\q\q(DT a)(JJ new)(NN Constitution))\\
\q\q\q\q\q\q\q (PP\\
\q\q\q\q\q\q\q\q\q\q(IN for)\\
\q\q\q\q\q\q\q (NP\\
\q\q\q\q\q\q\q\q\q\q(NP\\
\q\q\q\q\q\q\q\q\q\q\q\q(DT the)(NNP United)(NNP States))\\
\q\q\q\q\q\q\q\q\q\q (PP\\
\q\q\q\q\q\q\q\q\q\q\q\q(IN of)\\
\q\q\q\q\q\q\q\q\q\q\q\q(NP\\
\q\q\q\q\q\q\q\q\q\q\q\q\q(NNP America))))))))))))))\\
\end{minipage}
}
\caption{Formated textual representation of parse tree}
\end{subfigure}
\caption{Example parse tree}
\label{fig:Tree_ex1}
\end{figure}

\section{Parse Tree Features}
\label{sec:features}

The richness of the parsed representation introduces the possibility
of many different feature vectors.  Of the many possible feature
vectors that might be chosen, four are discussed here.  Examples are
provided based on the sentence above to illustrate the features.

\paragraph{All Subtrees}
One set of features is the set of all subtrees of a given depth
encountered among all the parsed sentences.  For example, Figure
\ref{fig:subtrees1} shows eleven subtrees of depth 3 extracted from
(\ref{eq:treeparse1}).  Subtrees of a given depth may appear more than
once within a sentence.  For example, the subtree
\begin{quote}
(NP(NP(DT)(JJ)(NN))(PP(IN)(NP(NP)(PP))))  
\end{quote}
appears twice in (\ref{eq:treeparse1}).

\begin{figure*}
  \centering
\begin{tabular}{cccc}
\begin{tabular}{>{\centering}p{0.2\linewidth}}
\scalebox{0.7}{
  \begin{forest}
[S[S[VP[VB][NP]]][NP[PRP]][VP[VBP][VP[VBN][PP][S]]]]
\end{forest}
} \\
{\scriptsize (S(S(VP(VB)(NP)))(NP(PRP9)) (VP(VBP)(VP(VBN)(PP)(S))))}
\end{tabular}
& 
\begin{tabular}{>{\centering}p{0.2\linewidth}}
\scalebox{0.7}{
\begin{forest}
[S[VP[VB][NP[NP][PP]]]]
\end{forest}
  } \\
  {\scriptsize   (S(VP(VB)(NP(NP)(PP))))  }
\end{tabular}  
& 
\begin{tabular}{>{\centering}p{0.2\linewidth}}
\scalebox{0.7}{
\begin{forest}
[VP[VB][NP[NP[DT][JJ][NN]][PP[IN][NP]]]]     
\end{forest}
  } \\
{\scriptsize (VP(VB)(NP(NP(DT)(JJ) (NN))(PP(IN)(NP))))  }
\end{tabular}
& 
\begin{tabular}{>{\centering}p{0.2\linewidth}}
\scalebox{0.7}{
\begin{forest}
[NP[NP[DT][JJ][NN]] [PP[IN][NP[NP][PP]]]]     
\end{forest}
  } \\
  {\scriptsize   (NP(NP(DT)(JJ)(NN))(PP(IN) (NP(NP)(PP))))   }
\end{tabular}
  \\[4em]
\begin{tabular}{>{\centering}p{0.2\linewidth}}
\scalebox{0.7}{
\begin{forest}
[PP[IN][NP[NP[DT] [NN]][PP[IN][NP]]]]     
\end{forest}
  } \\
  {\scriptsize  (PP(IN)(NP(NP(DT)(NN)) (PP(IN)(NP))))     }
\end{tabular}
& 
\begin{tabular}{>{\centering}p{0.2\linewidth}}
\scalebox{0.7}{
\begin{forest}
[NP[NP[DT][NN]][PP[IN][NP[DT][VBG][JJ][NN]]]]     
\end{forest}
  } \\
  {\scriptsize  (NP(NP(DT)(NN))(PP(IN) (NP(DT)(VBG)(JJ)(NN))))  }
\end{tabular}
& 
\begin{tabular}{>{\centering}p{0.2\linewidth}}
\scalebox{0.7}{
\begin{forest}
[VP[VBP][VP[VBN][PP[IN]][S[VP]]]]     
\end{forest}
  } \\
  {\scriptsize   (VP(VBP)(VP(VBN) (PP(IN))(S(VP))))     }
\end{tabular}
& 
\begin{tabular}{>{\centering}p{0.2\linewidth}}
\scalebox{0.7}{
\begin{forest}
[VP[VBN][PP[IN]][S[VP[TO][VP]]]]     
\end{forest}
  } \\
  {\scriptsize   (VP(VBN)(PP(IN)) (S(VP(TO)(VP))))     }
\end{tabular}
\\[4em]
\begin{tabular}{>{\centering}p{0.2\linewidth}}
\scalebox{0.7}{
\begin{forest}
[S[VP[TO][VP[VB][PP]]]]     
\end{forest}
  } \\
  {\scriptsize   (S(VP(TO)(VP(VB)(PP))))     }
\end{tabular}
& 
\begin{tabular}{>{\centering}p{0.2\linewidth}}
\scalebox{0.7}{
\begin{forest}
[VP[TO][VP[VB][PP[IN][NP]]]]     
\end{forest}
  } \\
  {\scriptsize  (VP(TO)(VP(VB)(PP(IN)(NP))))     }
\end{tabular}
& 
\begin{tabular}{>{\centering}p{0.2\linewidth}}
\scalebox{0.7}{
\begin{forest}
[VP[VB][PP[IN][NP[NP][PP]]]]     
\end{forest}
} \\
  {\scriptsize (VP(VB)(PP(IN)(NP(NP)(PP))))     }
\end{tabular}
& 
\begin{tabular}{>{\centering}p{0.2\linewidth}}
\scalebox{0.7}{
\begin{forest}
[PP[IN][NP[NP[DT][JJ][NN]][PP[IN][NP]]]]     
\end{forest}
} \\
  {\scriptsize (PP(IN)(NP(NP(DT)(JJ) (NN))(PP(IN)(NP))))     }
\end{tabular}
\\[4em]
\begin{tabular}{>{\centering}p{0.2\linewidth}}
\scalebox{0.7}{
\begin{forest}
[NP[NP[DT][JJ][NN]][PP[IN][NP[NP][PP]]]]     
\end{forest}
  } \\
  {\scriptsize   (NP(NP(DT)(JJ)(NN)) (PP(IN)(NP(NP)(PP))))     }
\end{tabular}
& 
\begin{tabular}{>{\centering}p{0.2\linewidth}}
\scalebox{0.7}{
\begin{forest}
[PP[IN][NP[NP[DT][NNP][NNP]][PP[IN][NP]]]]     
\end{forest}
  } \\
  {\scriptsize   (PP(IN)(NP(NP(DT) (NNP)(NNP))(PP(IN)(NP))))    }
\end{tabular}
& 
\begin{tabular}{>{\centering}p{0.2\linewidth}}
\scalebox{0.7}{
\begin{forest}
[NP[NP[DT][NNP][NNP]][PP[IN][NP[NNP]]]]     
\end{forest}
  } \\
  {\scriptsize  (NP(NP(DT)(NNP) (NNP))(PP(IN)(NP(NNP))))     }
\end{tabular}
\end{tabular}
\caption{Some subtrees of depth 3 extracted from the tree in
  (\ref{eq:treeparse1})}
  \label{fig:subtrees1}
\end{figure*}

Across all the sentences in the documents considered, there is a very
large number of subtrees.  This leads to vectors of very high
dimension.  This is a problem that is dealt with later.

\begin{figure}[h]
\begin{center}
\begin{tabular}{ccc}
\scalebox{0.8} {
\begin{minipage}{0.2\textwidth}
\begin{forest}
[S[PP][NP][VP]]
\end{forest}
\end{minipage}
}
& 
\scalebox{0.8}{
\begin{minipage}{0.3\textwidth}
\begin{forest}
[S[PP[IN][NP]][NP[PRP]][VP[VBP][VP]]]
\end{forest}
\end{minipage}
}
&
\scalebox{0.8}{
\begin{minipage}{0.49\textwidth}
\begin{forest}
[S[PP[IN][NP[NP][PP]]][NP[PRP]][VP[VBP][VP[VBN][PP][S]]]]
\end{forest} 
\end{minipage}
}    
\\
{\tiny  (S(PP)(NP)(VP))} &
{\tiny  (S(PP(IN)(NP))(NP(PRP))(VP(VBP)(VP))) } &
{\tiny  (S(PP(IN)(NP(NP)(PP)))(NP(PRP))(VP(VBP)(VP(VBN)(PP)(S)))) } \\
  One level & Two levels & Three levels
\end{tabular}
\end{center}
\caption{Rooted Subtrees of the tree in (\ref{eq:treeparse1}) of one,
  two, and three levels}
\label{fig:treelevels}
\end{figure}
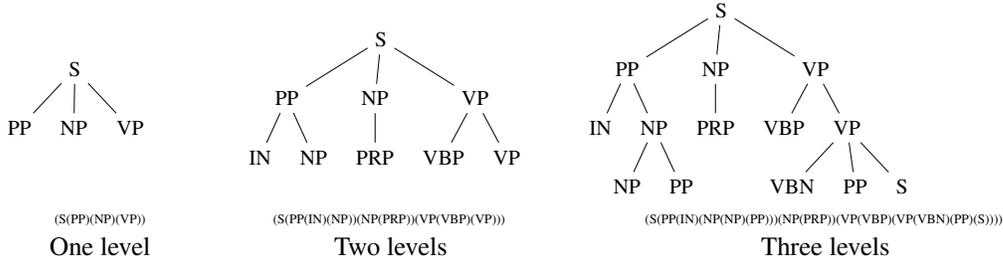

\paragraph{Rooted Subtrees}
A rooted subtree is a subtree of a tree whose root node is the root
node of the overall tree, down to some specified level.  The first few
rooted subtrees can be thought of summarizing the general structure of
a sentence, with the amount of detail in the summary related to the
number of levels of the subtree.  Fig. \ref{fig:treelevels}
illustrates the subtrees of levels one, two, and three for the tree of
Fig. \ref{fig:Tree_ex1}.

\paragraph{Part-of-Speech} A simple set of features ignores the tree
structure, and simply extracts the counts of tokens in the parse tree.
For (\ref{eq:treeparse1}), the counts of the POS are
\begin{center}
\begin{tabular}{cccccccccccccc}
  S & P & IN & NP & DT & JJ & NN & PRP & VP & VBP &  VBN & TO & VB & NNP \\
  2  & 7 & 7  & 11 & 5  & 4 & 4 &     1 &     4  &  1    & 1      & 1   & 1   &  3
\end{tabular}
\end{center}

\paragraph{POS by Level}
A more complicated set of features is the histogram of tokens at each level of the tree.  For the tree of (\ref{eq:treeparse1}), this is shown in Table \ref{tab:histlevelpos}.

\begin{table}
  \captionof{table}{POS counts by level for the tree (\ref{eq:treeparse1}).}
  \centering
\begin{tabular}{l|lllllll}
Level & Counts \\
1       &  S: 1 \\
2       &  PP: 1 & NP: 1 & VP : 1 \\
3       & IN : 1 & NP: 1 & PRP: 1 & VBP: 1 & VP: 1 \\
4       & NP: 1 & PP: 1 & VBN: 1 & PP: 1 & S: 1 \\
5       & DT: 1 & JJ: 1 & NN: 1 & IN: 1 & NP: 1 & IN: 1 & VP: 1 \\
6       & NP: 1 & PP: 1 & TO: 1 & VP: 1 \\
7       & DT: 1 & NN: 1 & IN: 1 & NP: 1 & VB:1 & PP: 1 \\
8:      & DT: 1 & JJ: 2 & NN: 1 & IN: 1 & NP: 1 \\
9:      & NP: 1 & PP: 1 \\
10:    & DT:1 & JJ: 1 & NN: 1 & IN: 1 & NP: 1 \\
11:    & NP: 1 & PP: 1 \\
12:    & DT: 1 & NNP: 1 & IN: 1 & NP: 1 \\
13:    & NNP: 1
\end{tabular}
  \label{tab:histlevelpos}
\end{table}

For purposes of author classification, the idea, of course, is to see
how the patterns in the feature vectors obtained from the sentences of
one author compare with the patterns in the feature vectors of other
authors.

\section{Classifier}
\label{sec:classifier}

This section describes the basic operation of the classifier employed
in the tests for this paper.  In this paper, when ``classes'' are
referred to, it refers to the different authors under consideration.
Let $k$ denote the number of classes (authors).

Let $n_i$ denote the number of documents associated with author $i$,
$i=1,2,\ldots, k$.  Let $\vbf_{i,j} \in \Rbb^m$ denote a feature
vector associated with a document (e.g. a normalized histogram of the
all subtrees counts for a document).  The set of all feature vectors
for author $i$ is formed by the columns the $\matsize{m}{n_i}$  matrix
$V_i = \begin{bmatrix}\vbf_{i,1} & \vbf_{i,2} & \ldots& \vbf_{i,n_i}
\end{bmatrix}$, $i=1,2,\ldots, k$.

An overall $\matsize{m}{n}$ data matrix is formed as
$V = \begin{bmatrix} V_1 & V_2 & \ldots & V_k \end{bmatrix}$, where
$\sum_{i=1}^{k} n_i = n$.  Let $N_i$ denote the set of column indices
of $V$ associated with the vectors in class $i$.  The centroids (that
is, the mean of the set of vectors) of the feature vectors for each
class are computed by
\[ \cbf^{(i)} = \frac{1}{n_i} \sum_{j=1}^{n_i} \vbf_{i,j} 
\text{ and the overall centroid is }
\cbf = \frac{1}{n} \sum_{i=1}^k \sum_{j=1}^{n_i} \vbf_{i,j}.
\]

In the tests performed for the investigation in this paper, the
classifier works as follows (see figure \ref{fig:statdepict1}).  
\begin{itemize}
\item For each feature vector under consideration
  $\vbf = \vbf_{i,j} \in V_i$ coming from class $i$, the vector
  $\vbf_{i,j}$ is removed from the pool of vectors in $V_i$, producing
  a set of vectors $\Vtilde_i$ and the centroid $\cbftilde_i$ of the
  resulting data is computed:
\[ \cbftilde^{(i)} = \frac{1}{n_i-1} \sum_{i \in N_i \setminus j}
V_i.
\]
Centroids for all the other classes are computed, but without removing
the vector under consideration, so $\cbftilde^{(j)} = \cbf^{(j)}$.

\item The vector $\vbf$ under consideration is compared with the class
  centroid for each class, and the estimated class is that class whose
  centroid is closest to $\vbf$, where the distance measure is simply
  Euclidean distance:
\[ \ihat = \argmin_j \| \vbf - \cbftilde^{(j)} \|
\]
\item A count of the vectors $\vbf$ which do not classify correctly is
  formed, where there are $n-1$ possible errors.
\end{itemize}

\section{Dimension Reduction}
\label{sec:reducedim}

As described in section \ref{sec:features}, the number of elements $m$
of the feature vectors can be very large.  It has been found to be
helpful to reduce the dimensionality of the feature vectors by
projecting them into a lower dimensional space.  The reduction of
dimension is similar to principle component analysis (PCA)
\cite{Jolliffe1986}, but is used when the dimension of the vectors
exceeds than the number of observations of vectors in the classes.
This has been used in other textual analysis problems
\cite{HowlandJeonPark,MoonHowlandGunther2006} and facial recognition
problems \cite{Howland2006}.  
(In \cite{ZhangWuNiuDing2014}, dimension reduction is accomplished in
a two-stage process, with PCA being following by a process similar to
the one described here.)  In this section we introduce the criterion
used to perform the projection.  In Appendix B, a few more details are
provided (see \cite{HowlandJeonPark,Howland2006} for more detail).

\begin{figure}
      \centering
\resizebox{0.5\textwidth}{!}{\input{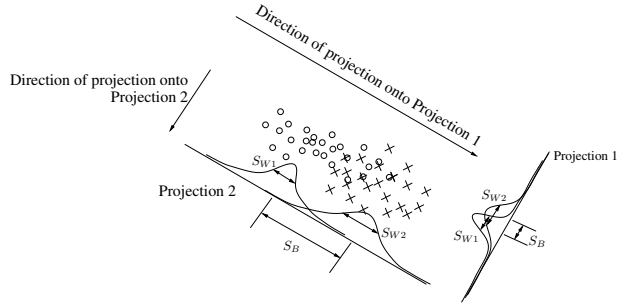}}
      \caption{Illustration of within-cluster and between cluster
        scattering and projection.}
      \label{fig:project1}
\end{figure}

While the feature vectors are in high-dimensional space, the salient
concepts of dimension reduction can be illustrated in low dimensional
space, such as figure \ref{fig:project1}.  In that figure there are
two 2-dimensional data sets, denoted with $\circ$ and $\times$,
respectively.  The problem is to determine for a given vector which
class it belongs to.  Also shown in the figure are two axes upon which
the data are projected.  (For ``projection,'' think of the shadow cast
by the data points by a light source high above the projection line.)
The 1-dimensional data produced in Projection 1 have a cluster widths
denoted by $S_{W2}$ and $S_{W2}$.  This is the within-cluster scatter,
a measure of the variance (or width) of the densities.  There is also
a between-cluster scatter, a measure of how far the cluster centroids
are from the overall centroid.  In Projection 1, the between-cluster
measure is rather small compared with the width of the cluster widths.
By contrast, the 1-dimensional data produced in Projection 2 have a
much larger between-cluster measure $S_B$.  The within-cluster scatter
$S_{W1}$ and $S_{W2}$ are also larger, but the between-cluster measure
appears to have grown more than these within-cluster measures.
Projection 2 produces data that would be more easily classified than
Projection 1.

More generally, one can conceive of rotating the data at various angle
with respect to the axis that data are projected upon.  At some
angles, the between cluster scatter will be larger compared to the
within-cluster scatters.

In light of this discussion, the goal of the projection operation is
to determine the best ``angles of rotation'' to project upon which
maximize the between-cluster scatter while minimizing the
within-cluster scatter.  In general, there are $k$ clusters of data to
deal with (not just the two portrayed in figure \ref{fig:project1}).
All this takes place in very high dimensional space, where we cannot
visualize the data, so this is done via mathematical transformations.
In higher dimensions, it is also not merely a matter of projecting
onto a single direction.  In $m$ dimensions, the dimension of the
projected data could be 1-dimension, 2-dimensions, etc., up to $m-1$
dimensions.  It is not known in advance what the best dimensionality
to project onto is, so this is one of the parameters examined in the
experiments described below.

With this discussion in place, we now describe the mathematics of how
the projection is accomplished.  For class $i$, with within-cluster
scatter matrix ---- that is, a measure of how the data in the in the
class vary around the centroid of the class --- is
\[ S_{Wi} = \sum_{j  \in N_i} (\vbf_j - \cbftilde^{(i)}) (\vbf_j -
  \cbftilde^{(i)})^T
\]
The total within-cluster scatter matrix is the sum of the within-class
scatter matrices,
\[ S_w = \sum_{i=1}^k \sum_{j \in N_i} (\vbf_j - \cbftilde^{(i)})(\vbf_j - \cbftilde^{(i)})^T.
\]
For the data considered here, which has high dimensions and not a lot
of training data, $S_w$ is singular, that is, not invertible.

The between-cluster scatter matrix is the scatter of the individual class centroids compared with the overall centroid,
\[ S_b = \sum_{i=1}^k \sum_{j \in N_i} (\cbftilde^{(i)} - \cbf) (\cbftilde^{(i)} - \cbf)^T =
\sum_{i=1}^k n_i  (\cbftilde^{(i)} - \cbf) (\cbftilde^{(i)} - \cbf)^T.
\]
The idea behind dimension reduction is to find a $\matsize{\ell}{m}$
matrix $G^T$ with $\ell < m$ then use $G^T$ to transform the data
according to  $\vbf_i^\ell = G^T \vbf_i$.
One may think of the matrix $G^T$ as providing rotation of the vectors
and selection of the dimensions which are retained after rotation.

The operation $\vbf_i^\ell = G^T \vbf_i$ may be thought of (naively)
as feature selection: elements of $\vbf_i$ are retained in
$\vbf_i^\ell$ which improve the clustering.  Actually, beyond mere
feature selection, the transformation $G^T$ also produces linear
combinations of feature vectors which improve the clustering of the
data (and hence may improve the classifier capability).

Based on the discussion above, the matrix $G^T$ is selected to
minimize the within-cluster scatter $S_w$ of the transformed data
while also making the between-cluster scatter $S_b$ as large as
possible.

It may be surprising that projecting into lower dimensional spaces can
improve the performance --- it seems to be throwing away information
that may be useful in the classification.  What happens, however, is
that the information discarded is in directions that are noisy, or may
be confusing to the classifier.  As the results below indicate,
projecting onto lower dimensions can significantly improve the classifier.


\section{The Federalist Papers}

With this background, we now turn attention to applying features
derived from statistical parsing to two different sets of documents,
the first being \emph{The Federalist Papers}.  {\em The Federalist
  Papers} consists of a series of 85 papers written around 1787 by
Alexander Hamilton, James Madison, and John Jay in support of the
U.S. Federal Constitution \cite{Fedpap}.  Of these papers, 51 are
attributed unambiguously to Hamilton, 14 to Madison, 5 to Jay, and 3
to both Hamilton and Madison.  The remaining twelve papers are of
uncertain attribution, but are known to be by either Madison or
Hamilton. In \cite{MostellerWallace1964,JanusJagenauer2005},
statistical techniques were used to determine that all twelve
ambiguous papers were due to Madison.

The authors used in this study are $\{$Hamilton, Madison, Jay,
UncertainHorM, HandM$\}$.  A machine-readable copy of {\em The
  Federalist Papers} was obtained from the Gutenberg project
\cite{fedascii}.  Each of the 85 papers is considered a separate
document, each of which is parsed into a separate file.
Below we consider the performance of classifiers based on the feature
vectors described above.

\paragraph{All Subtrees} Subtree extraction is done on all trees for
each of the \emph{Federalist} papers.  Table \ref{tab:fednums1} shows
the total number of sentences and words for each author considered.
The table also shows the number of different subtrees at any level for
the depths considered.  The number of subtrees grows rapidly with the
depth, leading to large feature vectors.  Table \ref{tab:fednums2}
shows the number of subtrees in the union of the subtrees of all the
authors and in the intersection of the subtrees of all the authors.


\begin{table} \RawFloats
\begin{minipage}{0.6\textwidth}
\caption{Number of subtrees in union and intersection of sets of all
    subtrees \label{tab:fednums1}}
\resizebox{\textwidth}{!}{\begin{tabular}{|l|l|p{.7in}|p{.4in}|p{.6in}|p{.6in}|p{.6in}|}
\hline
    Author & \# Docs & Total~\#  \hbox{Sentences} & Total \# Words &
                                                                  \# Subtrees of depth 2 &                                                                  \# Subtrees of depth 3 &   \# Subtrees of depth 4  \\
    \hline 
Hamilton    &     51   & 3126 & 206586 & 9660 & 19785 & 25834\\
Madison      &    14   & 1034 & 65492 & 4653 & 8182 & 9419 \\
Jay               &    5     & 159  & 11732 & 1590 & 2150 & 2044\\
UncertainHorM &  12& 688 & 40607 & 3200 & 5363 & 6077 \\
    HandM        &  3       & 154  & 7928 & 1007 & 1373 & 1293 \\
    \hline
\textbf{Totals} & 85 & 5161 &  332345 & 20110 & 36853 & 44667 \\
\hline
                   \end{tabular}
                   }
\end{minipage}
\hfill
%
%
\begin{minipage}{0.3\textwidth}
\caption{Summary statistics of \emph{Federalist} data\label{tab:fednums2} }
\resizebox{0.8\textwidth}{!}{
\begin{tabular}{|l|l|l|l|}
\hline
Depth & \# in union & \# in intersection \\
\hline
2   & 14377 & 283 \\     
3   & 30121 & 195 \\
4    & 39607 & 78 \\
\hline    
\end{tabular}
}
\end{minipage}
\end{table}

To form feature vectors using the subtree information, the top $N$
subtrees (by count) for each author are selected (where we examined
$N=5, 10, 20,$ and $30$), then the union across authors was formed of
these top subtrees.  In the tables below, the number of subtrees in
the union of the top $N$ is denoted as ``length(union)''.  The fact
that this number exceeds $N$ indicates that not all authors have the
same top $N$ subtrees.  This length(union) is $m$, the dimension of
the feature vector used for classification before projection into a
lower dimensional space.

The subtrees in the union of the top $N$ form the row labels in a
term-by-document matrix, where the terms (rows) are the subtrees and
there is one column for each paper.  This term-by-document is filled
with counts for each subtree, then the term-by-document matrix is
normalized so that each column sum is equal to 1.  Classification was
done by nearest distance to the class centroid, as described in
section \ref{sec:classifier}.

\begin{figure}\RawFloats
  \captionsetup[table]{position=top}
\begin{minipage}{0.45\textwidth}
\captionof{table}{Classification of \emph{Federalist} papers based on
  ``all subtree'' feature vectors}
  \label{tab:fedsubtree1}
  \centering
\resizebox{\textwidth}{!} {
\begin{tabular}{|l|l|l|l|llll|}
\hline
  top  $N$ & \shortstack{subtree \\ depth}
                   & length(union)  & \# Err &  \# Err$_2$& \# Err$_3$& \# Err$_4$& \# Err$_5$\\
  \hline
  5     &       2   &           16    &  25    &      29   &       10   &       13   &         15 \\
         &       3   &           45    &  22    &      12   &         5   &        2    &         3 \\
         &       4   &          103   &  25    &       1    &         1   &        2    &          2 \\
  \hline
  10   &       2   &        36       & 21   &    27      &         5   &        4    &         6 \\
         &       3   &        80       & 20     &     1     &         1   &        1    &         2 \\
         &       4   &      194       & 22    &     1     &         1   &        1    &         2 \\
  \hline
  20   &       2   &       82        &  16   & 1        &         1   &        1    &    2 \\
         &       3   &       182      &  19       & 1        &        1   &         1   &    2 \\
         &       4   &      402       & 20      &  1       & 1           &         1   & 2 \\
\hline
 30    &       2   &     146        & 15     &  1      &  1          &         1   & 2 \\
         &       3   &     286        & 19      &  1       & 1           & 1            & 2 \\
         &       4   &     700        & 23      &   1      &  1          & 1            & 2 \\
  \hline
\end{tabular}
}
\end{minipage}
\qquad \qquad 
\begin{minipage}{0.45\textwidth}
  \centering
  \includegraphics[width=\textwidth]{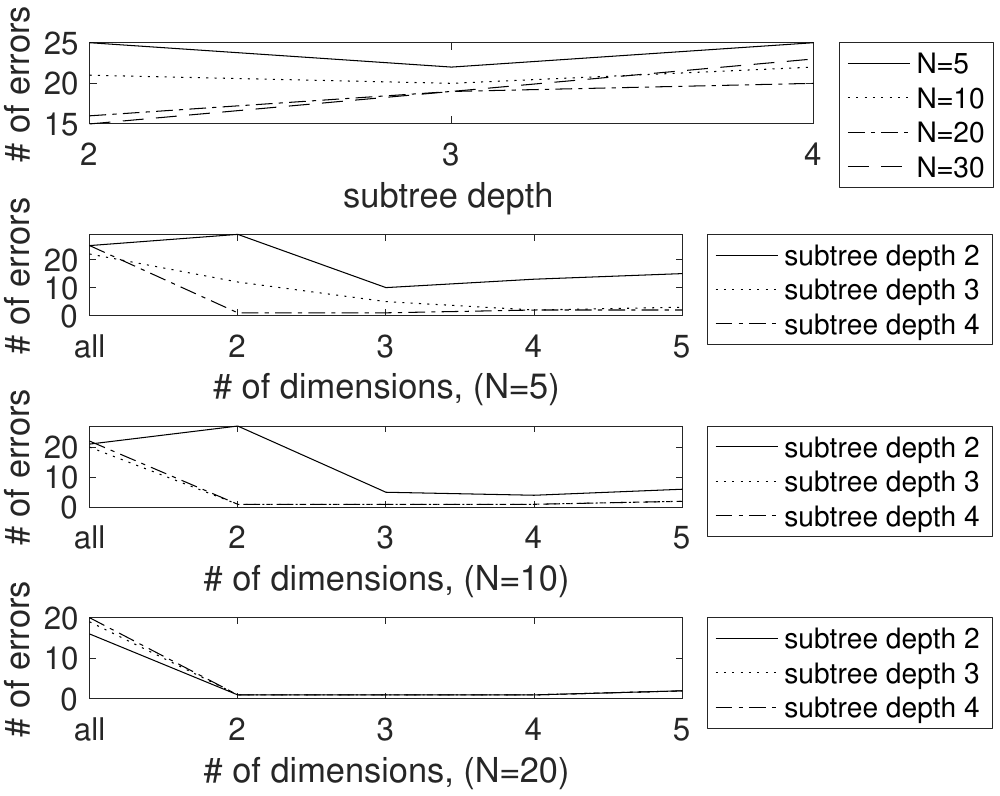}
  \caption{Error counts for the ``all subtree" features}
  \label{fig:feddata1}
\end{minipage}
\end{figure}

Classification results are shown in Table \ref{tab:fedsubtree1}.  The
results (most of them) are also shown in figure \ref{fig:feddata1}.
The test conditions are the number $N$ (for the top $N$), the depth of
the tree, and the dimension of the reduced dimension space $\ell$.
There is an error count in the column ``\# Err'', which is the number
of errors (out of 85) using the original high-dimensional feature
vectors.  There are also error counts \# Err$_{\ell}$, for
$\ell = 1, 2, 3, 4, 5$, which are the number of errors for data
projected in the $\ell$-dimensional space as described in section
\ref{sec:reducedim}.  The \# Err column never achieves a value less
than 16, illustrating that the raw subtree features do not provide
good classification.  However, the reduced-dimension data can achieve
good classification.  For example, with top $N=5$, subtrees of depth 4
achieve an error count of 1 for $\ell=2$ and $\ell=3$ dimensional
projection.  In fact, the error count is actually better than the
table shows.  For all error counts of 1, the one error that occurs is
a classification of the author HandM as the author Madison.  Since it
is understood that the HandM papers were actually written by Madison,
this is a correct classification.

There is clearly a broad ``sweet spot'' for these features.  Taking
the top $N$ at least 10, a subtree depth of 3 or 4, and projected
dimension of $\ell=2, 3$ or 4 provides the best performance.
Interestingly, in all cases, moving to $\ell=5$ actually slightly
increases the number of errors to 2.

These results indicate that the subtree feature \emph{does} provide
information which can distinguish authorship, with appropriate
weighting of features and selection of the dimension.

\paragraph{Rooted Subtrees}
We next considered using rooted subtrees as feature vectors.  A few
examples of these trees created from a Hamilton paper in the
\emph{The Federalist Papers} are shown in figure \ref{fig:exroottrees1}.
There is quite a variety of possibilities (more than initially
anticipated).

Some summary information about the number of different trees by author
and level is shown in Table \ref{tab:fedrootedsum}.  Table
\ref{tab:fedrootedsum2} shows the number of rooted trees in the union
of the trees across the authors, and the number of trees common to all
authors (the intersection).  There is so much variety in these trees
that it is only at level 1 that there are any trees common to all
authors, which is the tree S(NP)(VP), that is, a sentence with a noun
phrase and a verb phrase.

The feature vectors were formed as follows.  For each level of rooted
subtree, the top $N$ trees for each author were selected, and the
union of these trees across the documents formed the terms in a
term-by-document matrix.  The number of trees obtained is shown in the
column \# Trees of Table \ref{tab:fedrootedsubtree1}.

Incidence counts were formed for each document of each user.
Classification was done by nearest distance to the class centroid, as
described in section \ref{sec:classifier}.  This gave rather high
error counts for each of the different levels.  Then the data was
projected into $\ell$-dimensional space, and the error counts \#
Err$_{\ell}$ are computed.  The results are shown in Table
\ref{tab:fedrootedsubtree1}.  The columns \# Err, \#Err$_2$, \ldots
\#Err$_5$ show the number of errors for full dimension then 2-,
\ldots, 5-dimensional projections.  Figure \ref{fig:feddata2}
graphically portrays the tabulated results.

As in the ``all subtree'' feature case, these low-dimensional
projections do very well.  In fact, as before, the error that occurs
in the case that there is one error is when the HandM author was
classified as the Madison author, which is in fact a correct
classification.

The rooted subtrees features have a very broad sweet spot where good
classification occurs.  For dimensions $\ell = 2, 3$ or 4, and at
least 2 levels does very well.  As for the all subtrees features, in
this case also: for all error counts of 1, the one error that occurs
is a classification of the author HandM as the author Madison.  Since
it is understood that the HandM papers were actually written by
Madison, this is a correct classification, so all documents were
correctly classified.

\begin{figure}
\centering
\scalebox{0.6}{\begin{forest}[S[S][NP][VP]]\end{forest}} \quad
\scalebox{0.6}{\begin{forest}[S[NP][VP]]\end{forest}}\quad
\scalebox{0.6}{\begin{forest}[S[S][S]]\end{forest}}\quad
\scalebox{0.6}{\begin{forest}[S[CC][NP][VP]]\end{forest}}\quad
\scalebox{0.6}{\begin{forest}[S[S][S][S][S][CC][S]]\end{forest}}
\bigskip

\scalebox{0.6}{\begin{forest}[S[S[VP]][NP[PRP]][VP[MD][VP]]]\end{forest}} \quad
\scalebox{0.6}{\begin{forest}[S[NP[CD][NNS]][VP[MD][VP]]]\end{forest}}\quad
\scalebox{0.6}{\begin{forest}[S[S[VP]][NP[PRP]][VP[VBZ][ADJP][SBAR]]]\end{forest}} \quad
\scalebox{0.6}{\begin{forest}[S[S[NP][VP]][S[SBAR][ADVP][NP][VP]]]\end{forest}} \quad
\scalebox{0.6}{\begin{forest}[S[CC][NP[DT][NNS]][VP[VBP][VP]]]\end{forest}}
\bigskip

\scalebox{0.6}{\begin{forest}[S[S[VP[VB][S]]][NP[PRP]][VP[MD][VP[VB][NP][SBAR]]]]\end{forest}} \quad
\scalebox{0.6}{\begin{forest}[S[NP[CD][NNS]][VP[MD][VP[VB][PP][PP]]]]\end{forest}}\quad
\scalebox{0.6}{\begin{forest}[S[S[VP[VB][SBAR]]][NP[PRP]][VP[VBZ][ADJP[JJ]][SBAR[IN][S]]]]\end{forest}}\quad
\caption{Example rooted trees of 1 level, 2 levels, 3 levels}
\label{fig:exroottrees1}
\end{figure}
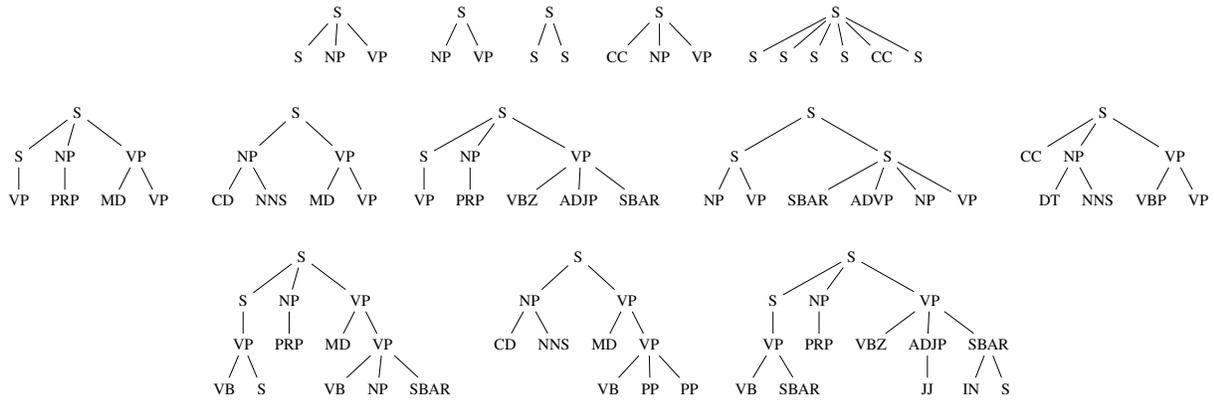

\begin{figure} \RawFloats
  \captionsetup[table]{position=top}
\begin{minipage}{0.6\textwidth}
\captionof{table}{Summary statistics of number of ``rooted subtrees'' of
    different levels}
  \label{tab:fedrootedsum}
\centering
\resizebox{\textwidth}{!}{
\begin{tabular}{|l|p{.6in}|p{.6in}|p{.6in}|p{.6in}|}
\hline
Author             & \# Rooted sub trees of 1 level  & \# Rooted subtrees of 2 levels & \# Rooted subtrees of 3 levels &  \# Rooted subtrees of 4 levels \\
\hline
Hamilton         & 285 & 1688                                 & 2732                                 & 2955 \\
Madison          & 156 & 703                                & 959                                 & 995 \\
Jay                   & 51 & 139                                   & 159                                   & 157 \\
UncertainHorM&111 &490                                     & 633                                  & 647 \\
  HandM            & 28  & 137                                  & 152                                   & 148 \\
  \hline
\end{tabular}
}
\end{minipage}
\qquad
\qquad
\begin{minipage}{0.3\textwidth}
  \captionsetup[table]{position=top}
\captionof{table}{Summary statistics of number of rooted trees unioned and
    intersected over authors}
\label{tab:fedrootedsum2}
  \centering
\resizebox{\textwidth}{!}{
\begin{tabular}{|l|p{.6in}|p{.6in}|}
\hline
  Level & \# in union & \# in intersection \\
\hline
1       & 398            & 1 \\
2       & 2625          & 0 \\
3       & 4425          & 0 \\
4       & 4808          & 0 \\
\hline
\end{tabular}
}
\end{minipage}
\end{figure}

\begin{figure}\RawFloats
\captionsetup[table]{position=top}
\begin{minipage}{0.45\textwidth}
\captionof{table}{Classification of \emph{Federalist} papers based on ``rooted subtree'' feature vectors}
\label{tab:fedrootedsubtree1}
\resizebox{\textwidth}{!}
{
\begin{tabular}{|l|l|l|l|llll|}
\hline
top $N$  &  level  & \# Trees   &  \# Err & \# Err$_2$ &  \# Err$_3$ &  \# Err$_4$ & \# Err$_5$  \\
\hline
5             &  1      &       37      &       36 &    15        & 13             &    11           &  2 \\
               &  2      & 168          &       34 &     1         &  1              &      1           & 2 \\
               & 3       & 371          &      35  &     1         &   1             & 1                &   2\\
               & 4       & 394          &      30  &    1          &  1              & 1                &  2 \\
\hline
10           & 1       & 140        &  33          & 1           & 1              & 1                 & 2 \\
               & 2       & 413        & 35           & 1           & 1              & 1                & 2 \\
               & 3       & 734        & 30           & 1           & 1              & 1                & 2 \\
               & 4       & 774        & 30           & 1            & 1              & 1                & 2 \\
\hline
20           & 1       & 323        &  33          & 1           & 1              & 1                 & 2 \\
               & 2       & 945        & 29           & 4           & 1              & 1                & 2 \\
               & 3       & 1462        & 31           & 1           & 1              & 1                & 2 \\
               & 4       & 1538        & 30           & 1            & 1              & 1                & 2 \\
30           & 1       & 381        &  34          & 1           & 1              & 1                 & 2 \\
               & 2       & 1416       & 29           & 1           & 1              & 1                & 2 \\
               & 3       & 2145        & 30           & 1           & 1              & 1                & 2 \\
               & 4       & 2283        & 30           & 1            & 1              & 1                & 2 \\
\hline
\end{tabular}               
}
\end{minipage}
\qquad \qquad
\begin{minipage}{0.45\textwidth}
  \centering
  \includegraphics[width=\textwidth]{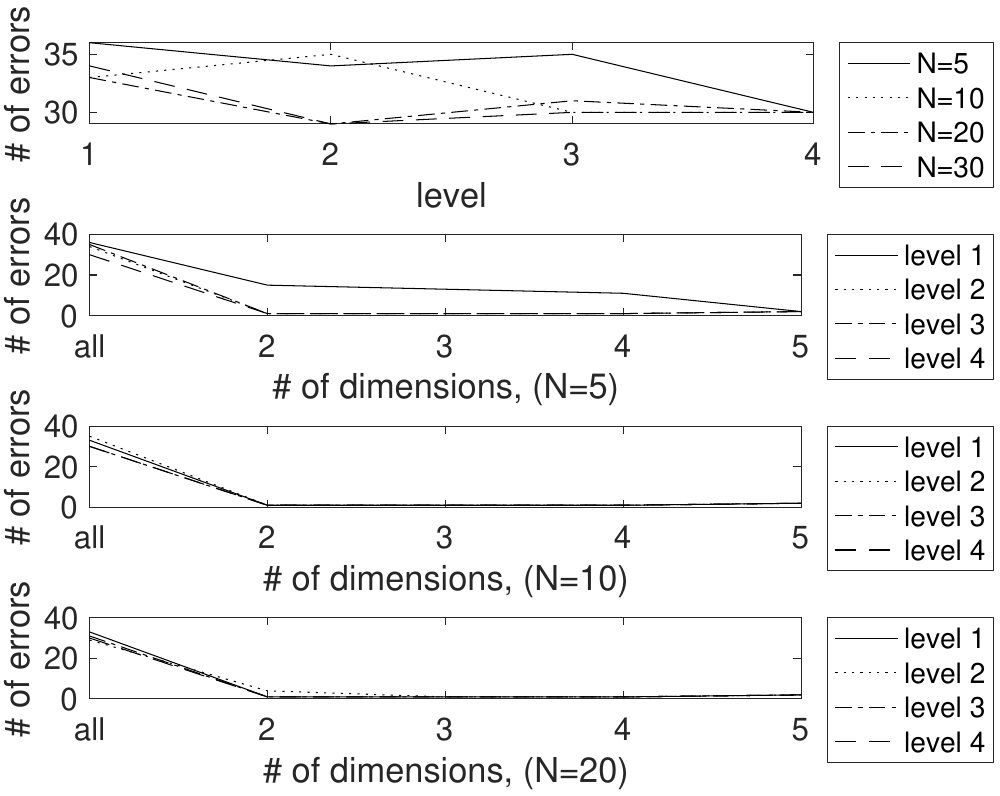}
  \caption{Error counts for the ``rooted subtree" features}
  \label{fig:feddata2}
\end{minipage}
\end{figure}

\paragraph{POS} POS is a seemingly natural way to classify documents,
but, contrary to expectations, it does not perform as well as the
tree-based features.  Feature vectors in this case are formed by
taking the top $N$ most common POS for each author, then forming the
union of these POS.  Feature vectors are formed by POS counts by
author and document, normalized.  Results are shown in Table
\ref{tab:fedPOS1}.  The raw error count for different values of $N$
are all greater than or equal to 23.  Moderate improvements can be
obtained by projecting the feature vectors to lower dimensional space,
with the errors for the $\ell$-dimensional projection denoted by \#
Err$_\ell$, for $\ell = 2, 3, 4, 5$.  Even in the best of
circumstances, the error counts is equal to 4.

We conclude that while the POS provides a measure of
distinguishability between authors, it does not provide the same
degree of distinguishability as that provided by the structural
information obtained from the parse trees.

\begin{table}
\caption{Classification of \emph{Federalist} papers based on POS vectors}
  \centering
\begin{tabular}{|l|l|l|llll|}
\hline
top $N$   & \# POS in union & \# Err & \# Err$_2$ &\# Err$_3$ &  \# Err$_4$ & \# Err$_5$  \\
\hline
5             & 7                        &  30      &      32       &       27       &    27             &  30  \\
10             & 17                    &  25      &      41       &       19       &    14             &  16  \\
20             & 28                    &  23      &      31       &       11       &    6             &  7  \\
30             & 46                    &  23      &      12       &       6       &    4             &  5  \\
\hline
\end{tabular}               
\label{tab:fedPOS1}
\end{table}

\begin{figure}\RawFloats
  \captionsetup[table]{position=top}
\begin{minipage}{0.45\textwidth}
\captionof{table}{Classification of \emph{Federalist} papers based on ``POS
    by level feature'' vectors\label{tab:POSbylevel1}}
  \centering
\resizebox{\textwidth}{!}{
\begin{tabular}{|l|l|l|l|llll|}
\hline
top $N$  &  level  & \# Trees   &  \# Err & \# Err$_2$ &  \# Err$_3$ &  \# Err$_4$ & \# Err$_5$  \\
\hline
5             & 1       & 9             &      42   &    11             & 35             &    39             &  40 \\
               & 2       & 14           &      35   &    31             &  30            &    24             &  28 \\
               & 3       & 12           &     46    &    38             &  31            &    32             &  35 \\
               & 4       & 11           &     49    &    33             &  37            &    39             &  39 \\
\hline
10           & 1       & 27           &      41   &    34             & 27             &    23             &  21 \\
               & 2       & 20           &      36   &    35             &  28            &    21             &  24 \\
               & 3       & 21           &     39    &    29             &  20            &    21             &  27 \\
               & 4       & 19           &     40    &    23             &  19            &    20             &  22 \\
\hline
20           & 1       & 33           &      41   &    36             & 18             &    18             &  18 \\
               & 2       & 44           &     33    &    16             &  5              &    5             &  6 \\
               & 3       & 36           &     38    &    23             &  13            &    9              &  9 \\
               & 4       & 40           &     39    &    34             &  12            &    10             &  11 \\
\hline
\end{tabular}               
}
\end{minipage}
\qquad \qquad
\begin{minipage}{0.45\textwidth}
  \centering
  \includegraphics[width=\textwidth]{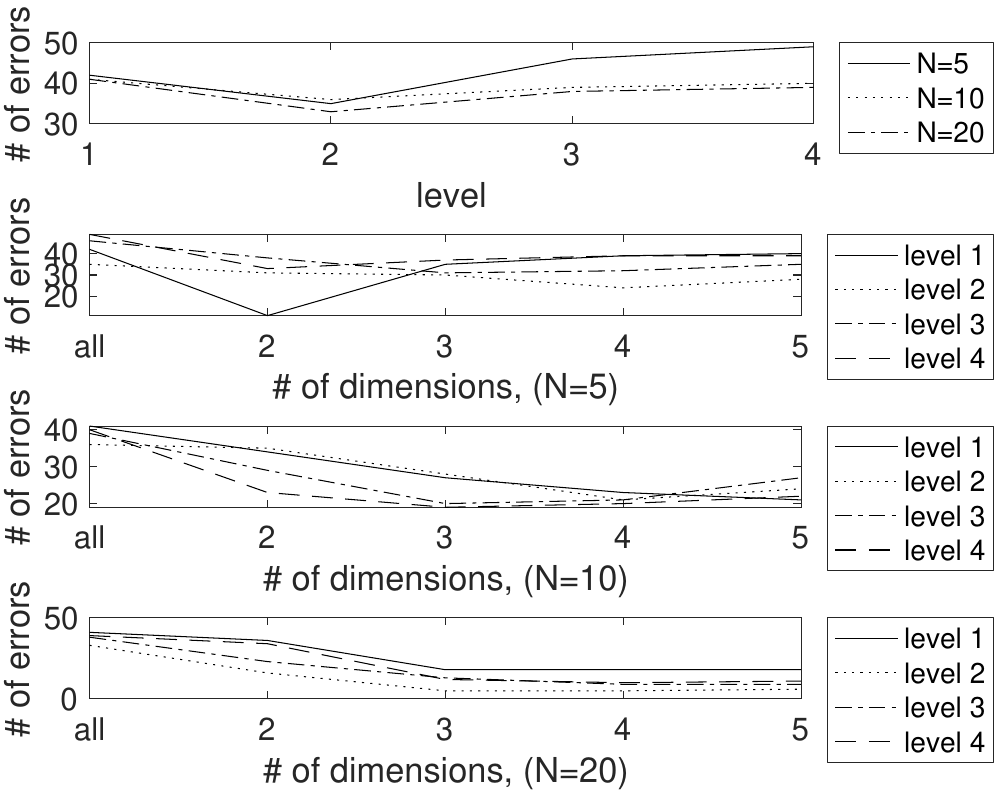}
  \caption{Error counts for the ``POS by level" feature vectors}
  \label{fig:feddata3}
\end{minipage}
\end{figure}

\paragraph{POS by Level} Table \ref{tab:POSbylevel1} shows the
classification results for feature vectors obtained using the POS by
level, using feature vectors formed in a manner similar to the other
features.  Figure \ref{fig:feddata3} provides a graphical
representation of this data.  This feature vector provides some
discrimination between authors, but fares substantially worse than the
purely tree-based features.

\subsection{Sanditon}

Up until shortly before her death in 1817, Jane Austen was working on
a novel posthumously titled {\em Sanditon} \cite[p. 20]{Poplawski}.
Before her death she completed a draft of twelve chapters (about
24,000 words).  The novel was posthumously ``completed'' by various
writers with varying success.  The version best known was published in
1975 \cite{Sanditon1975}, coathored by ``Another Lady,'' whose
identity remains unknown.  Whoever she was, she was a fan of Austen's
and attempted to mimic her style.  Of this version, it was said, it
``received, as compared with [its] predecessors, a warm reception from
the English critics.'' \cite[p. 76]{Hopkinson}.  Notwithstanding its
literary appeal and the attempts at imitating the conscious habits of
Austen, she failed in capturing the unconscious habits of detail:
stylometric analysis has been able to distinguish between the
different authors \cite[Chapter 16]{Morton1978}.

We obtained a computer-readable document from the Electronic Text
Center at the University of Virginia Library \cite{sandhtml}.  The
document was evidently obtained optical character recognition (OCR)
from scanned documents, so it was necessary to carefully spell-check
the document, but contemporary spellings were retained.  Two documents
were produced, the first for Austen (with 1176 sentences) and the
second for Other (with 2559 sentences).  These were split into
segments (for purposes of testing the classification capability).  The
Austen document had two segments of length 588 sentences.  The Other
document had four segments of lengths 640, 640, 640, 639.  Subtrees of
various depths were extracted from the segments, and these were
classified the same way as the \emph{Federalist} papers.  Summary
statistics about the documents are provided in Table
\ref{tab:sandnums1}.

Despite the attempt to duplicate Austen's style, the segments for the
different authors readily classify according to author, as shown below.

\begin{figure}\RawFloats
  \captionsetup[table]{position=top}
\begin{minipage}{0.6\textwidth}
\captionof{table}{Summary statistics of \emph{Sanditon} data}
  \label{tab:sandnums1}
  \centering
\resizebox{\textwidth}{!}{
  \begin{tabular}{|l|l|p{.7in}|p{.4in}|p{.6in}|p{.6in}|p{.6in}|}
\hline
    Author & \# Docs & Total~\#  \hbox{Sentences} & Total \# Words &
                                                                  \# Subtrees of depth 2 &                                                                  \# Subtrees of depth 3 &   \# Subtrees of depth 4  \\
    \hline 
Austen    &     2   & 1176  & 26342 & 4921 & 7378 & 7460\\
Other      &    4   & 2559 & 55453 & 8194 & 13795 & 15266 \\
    \hline
\textbf{Totals} & 6 & 5161 &  332345 & 20110 & 36853 & 44667 \\
\hline
  \end{tabular}
}
\end{minipage}
\qquad \qquad
\begin{minipage}{0.3\textwidth}
\captionof{table}{Number of subtrees in union and intersection of sets of subtrees for \emph{Sanditon}.}
  \label{tab:sandnums2}
  \centering
\resizebox{\textwidth}{!}{\begin{tabular}{|l|l|l|l|}
\hline
Depth & \# in union & \# in intersection \\
\hline
2   & 11189 & 1926 \\     
3   & 19277 & 1896 \\
4    & 21560 & 1166 \\
\hline    
\end{tabular}
}
\end{minipage}
\end{figure}



\begin{figure}[t]\RawFloats
  \captionsetup[table]{position=top}
\begin{minipage}{0.45\textwidth}
\captionof{table}{Classification of \emph{Sanditon} based on ``all subtrees''
    feature vectors \label{tab:sandsubtree1}}
\centering
\resizebox{\textwidth}{!}{
\begin{tabular}{|l|l|l|l|llll|}
\hline
  top  $N$ & \shortstack{subtree \\ depth}
                   & length(union)  & \# Err & \# Err$_2$& \# Err$_3$& \# Err$_4$& \# Err$_5$\\
\hline
5             &  2             & 9         &     0      &  0             &  2           & 4             &  5     \\
               &  3             & 11       &     0      &  0             & 2            & 4             & 5      \\
               &  4             & 15       &     2      & 0              & 2            & 5             & 5      \\
\hline
10           &  2             & 12       &     0      &  0             & 2           & 4             &  5     \\
               &  3             & 17       &     0      &  0             & 2           & 4             & 5      \\
               &  4             & 29       &     2      & 0              & 3            & 4             & 5      \\
\hline
20           &  2             & 30       &     0      &  0             & 2           & 4             &  5     \\
               &  3             & 35       &     0      &  0             & 1           & 3             & 5      \\
               &  4             & 58       &     1      & 0              & 2            & 4             & 5      \\
\hline
30           &  2             & 46       &     0      &  0             & 2           & 4             &  5     \\
               &  3             & 51       &     0      &  0             & 1           & 3             & 5      \\
               &  4             & 88       &     2      & 0              & 3            & 4             & 5      \\
\hline
\end{tabular}
}
\end{minipage}
\qquad \qquad
\begin{minipage}{0.45\textwidth}
  \centering
\includegraphics[width=\textwidth]{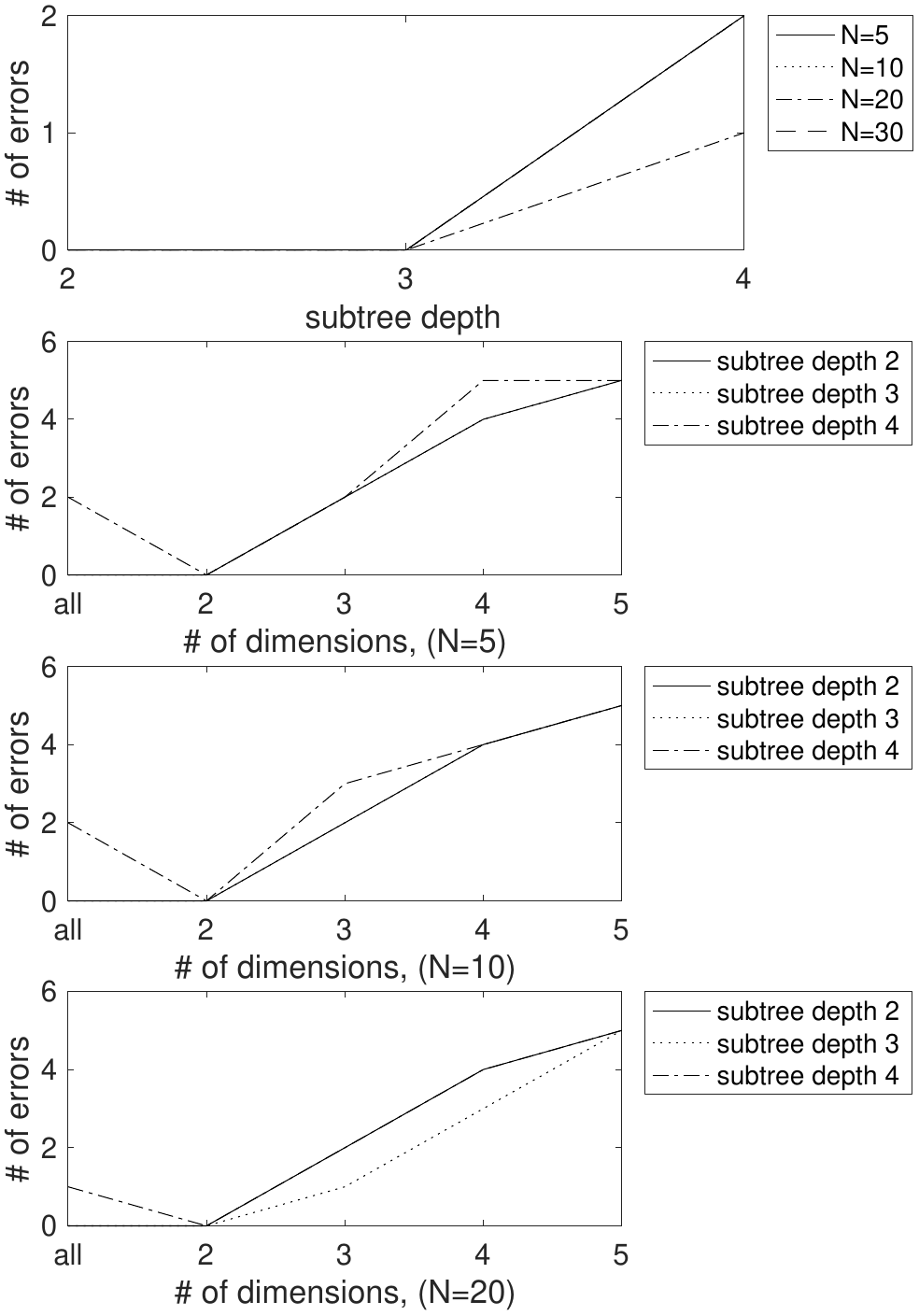}  
\caption{Classification of Sanditon based on ``all subtrees'' feature
  vectors}
\label{fig:sandidat1}
\end{minipage}
\end{figure}

\paragraph{All Subtrees} For each six of the documents (two Austen,
four Other), counts of all subtrees were extracted.  As for the
\emph{Federalist} papers, the top $N$ counts were extracted for
$N=5,10,20,30$, and the union of these features was formed.  This was
done for subtrees of depth 2, 3, and 4.  The number of trees in the
union and intersection of these sets is shown in Table
\ref{tab:sandnums2}.

Classifier results for the all subtrees feature are shown in Table
\ref{tab:sandsubtree1}, and also portrayed in figure
\ref{fig:sandidat1}.  As is shown, even with the full dimensionality
(without projecting into a lower dimensional space), separation can be
done completely accurately.  On the other hand, the projected feature
vectors do not generally perform as well as the full-dimensional
data.  This differs from how the lower dimensional projections worked
for the \emph{Federalist} documents.

\begin{figure}\RawFloats
  \captionsetup[table]{position=top}
\begin{minipage}{0.45\textwidth}
\captionof{table}{Classification of \emph{Sanditon} based on ``rooted subtree'' feature vectors}
\label{tab:sandrootedsubtree1}
\centering
\resizebox{\textwidth}{!}{
\begin{tabular}{|l|l|l|l|llll|}
\hline
top $N$  &  level  & \# Trees   &  \# Err & \# Err$_2$ &  \# Err$_3$ &  \# Err$_4$ & \# Err$_5$  \\
\hline
5             &  1      &      8          &      0   &        5        &       1             &    3            &  5 \\
               &  2      &    11          &      2    &       0         &      1             &    3            &  6 \\
               &  3      &    27          &      4    &       0         &      3             &    4            &  6 \\
               &  4       &   27          &      2    &       0         &      2             &    3            &  6 \\
\hline
10             &  1    &     15          &      0   &        0        &       1             &    3            &  5 \\
               &  2      &    28          &      4    &       0         &      3             &    3            &  5  \\
               &  3      &    56          &      3    &       0         &      2             &    4            &  5  \\
               &  4       &   56          &      2    &       0         &      2             &    4            &  5 \\
\hline
20             &  1    &     31          &      0   &        0        &       2             &    3            &  5 \\
               &  2      &    60          &      2    &       0         &      2             &    4            &  5 \\
               &  3      &    115          &      2    &       0         &      1             &    5            &  5 \\
               &  4       &   116          &      2    &       0         &      2             &    3            &  5 \\
\hline
30             &  1    &     59          &      0   &        0        &       2             &    3            &  5 \\
               &  2      &    106          &      2    &       0         &      2             &    3            &  5  \\
               &  3      &    174          &      2    &       0         &      2             &    5            &  5 \\
               &  4       &   176          &      2    &       0         &      2             &    3            &  5 \\
\hline
\end{tabular}               
}
\end{minipage}
\qquad\qquad
\begin{minipage}{0.45\textwidth}
 \centering
\includegraphics[width=\textwidth]{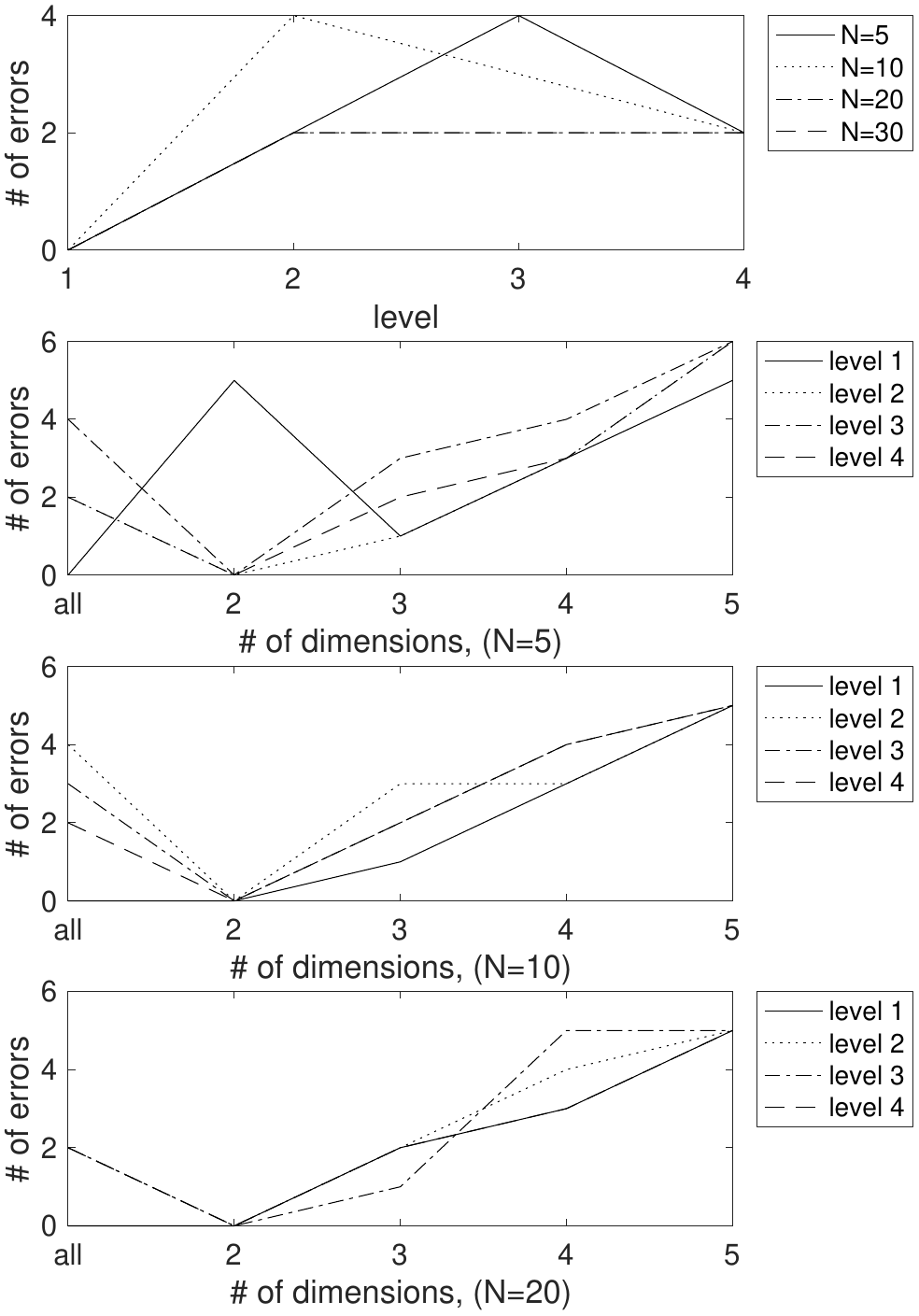}  
  \caption{Classification of Sanditon based on ``rooted subtree'' feature
    vectors}
  \label{fig:sandidat2}
\end{minipage}
\end{figure}

\paragraph{Rooted Subtrees}
We next considered using rooted subtrees as feature vectors.  Feature
vectors were formed in the same way as for the \emph{The Federalist
  Papers}.  Results are shown in Table \ref{tab:sandrootedsubtree1}
and portrayed in figure \ref{fig:sandidat2}.
While not as effective at distinguishing as the subtrees features,
this feature still shows the ability to distinguish between authors.

\paragraph{POS} POS feature vectors were extracted in the same manner
as for the \emph{The Federalist Papers}.  Data up to Err$_4$ were
produced.  The POS data was able to effectively distinguish between
authors, more effectively than for the \emph{The Federalist Papers}.
Reducing the dimensionality did not improve the classifier (and beyond
$\ell=2$ made it worse).

\begin{table}
\caption{Classification of \emph{Sanditon} based on POS vectors}
  \centering
\begin{tabular}{|l|l|l|lll|}
\hline
top $N$   & \# POS in union & \# Err & \# Err$_2$ &\# Err$_3$ &  \# Err$_4$   \\
\hline
5             & 5                        &  0        &      3          &    4           &    5      \\
10             & 11                    &  0      &        0          &       1        &    3  \\
20             & 21                    &  0      &        0          &       2        &    4    \\
30             & 36                    &  0     &         0          &       2       &    4    \\
\hline
\end{tabular}               
\label{tab:sandPOS}
\end{table}

\begin{figure}\RawFloats
  \captionsetup[table]{position=top}
\begin{minipage}{0.45\textwidth}
  \centering
\captionof{table}{Classification of \emph{Sanditon} based on ``POS by level'' feature vectors}
\label{tab:sandPOSbylevel1}
\resizebox{\textwidth}{!}
{
\begin{tabular}{|l|l|l|l|lll|}
\hline
top $N$  &  level  & \# Trees   &  \# Err & \# Err$_2$ &  \# Err$_3$ &  \# Err$_4$  \\
\hline
5             & 1       & 5             &      0   &    2             & 4             &    4     \\
               & 2       & 5             &      0   &    1             &  2            &    5      \\
               & 3       & 7             &     1    &    0             &  1            &    3       \\
               & 4       & 7             &     0    &    0             &  2            &    3        \\
\hline
10           & 1       & 11             &      0   &    0             & 2             &    4     \\
               & 2       & 11             &      0   &    0             &  2            &    4      \\
               & 3       & 13             &     1    &    0             &  2            &    3       \\
               & 4       & 10             &     0    &    0             &  2            &    3        \\
\hline
20           & 1       & 26             &      0   &    0             & 3             &    4     \\
               & 2       & 21             &      0   &    0             &  2            &    4      \\
               & 3       & 21             &     0    &    0             &  1            &    3       \\
               & 4       & 21             &     0    &    0             &  2            &    4        \\
\hline
30           & 1       & 42             &      0   &    0             & 3             &    4     \\
               & 2       & 36             &      0   &    0             &  2            &    4      \\
               & 3       & 35             &     0    &    0             &  1            &    3       \\
               & 4       & 33             &     0    &    0             &  2            &    4        \\
\hline
\end{tabular}               
}
\end{minipage}
\qquad \qquad
\begin{minipage}{0.45\textwidth}
 \centering
\includegraphics[width=\textwidth]{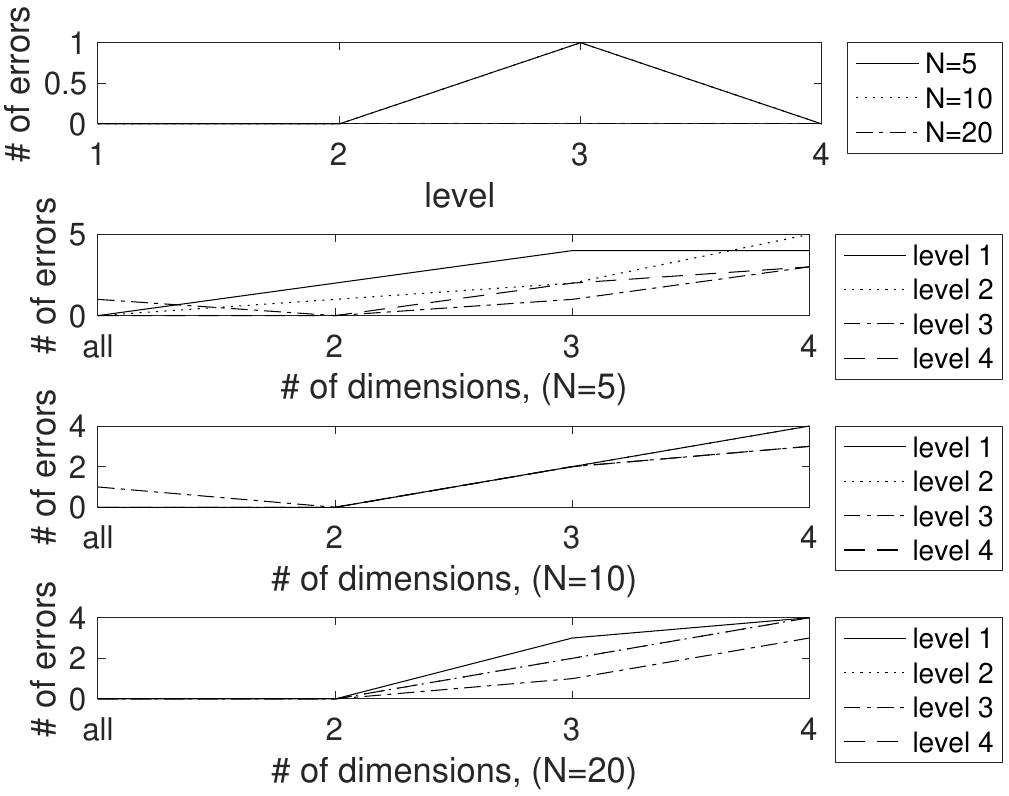}  
\caption{Classification of Sanditon based on ``POS by level'' feature
  vectors}
\label{fig:sandidat3}
\end{minipage}
\end{figure}

\paragraph{POS by Level} 
 POS by Level feature vectors were extracted in the
same manner as for the \emph{The Federalist Papers}.  Data up to Err$_4$
were produced.  The classification results are shown in Table
\ref{tab:sandPOSbylevel1} and portrayed in figure \ref{fig:sandidat3}.   

The POS by Level data was able to
effectively distinguish between authors, more effectively than for the
\emph{The Federalist Papers}.  Reducing the dimensionality did not improve
the classifier (and beyond $\ell=2$ made it worse).

\section{Conclusions, Discussion, and Future Work}

As this paper has demonstrated, information drawn from statistical
parsing of a text can be used to distinguish between between authors.
Different sets of features have been considered (all subtrees, rooted
subtrees, POS, and POS by Level), with different degrees of
performance among them.  Other than the POS these features have not
been previously considered (to the knowledge of the authors),
including in the large set of features examined in
\cite{ZhangWuNiuDing2014}.  This suggests that these tree-based
features, especially the features based on all subtrees, may be
beneficially included among other features.

It appears that the \emph{Sanditon} texts are easier to classify than
the \emph{The Federalist Papers}.  Even without the generally
performance-enhancing step of dimension reduction, \emph{Sanditon}
classifies well, even using the POS feature vectors which are not as
strong when applied to the \emph{The Federalist Papers}.  This is
amusing, since the completer of \emph{Sanditon} attempted to write in
an imitative style, suggesting that these structural features are not
easily faked.

The methods examined here does not preclude the excellent work on
author identification that has previously been done, which is usually
done using more obvious features in the document (such as word counts,
with words selected from some appropriate set).  This makes previous
methods easier to compute.  But at the same time, it may make it easy
to spoof the author identification.  The grammatical parsing provides
more subtle features which will be more difficult to spoof.

Another tradeoff is the amount of data needed to extract a
statistically meaningful feature vector.  The number of trees --- the
number of feature elements --- quickly becomes very large.  In order
to be statistically significant a feature element should have multiple
counts.  (Recall that for the chi-squared test in classical statistics
a rule of thumb is that at least five counts are needed.)  This need
to count a lot of features indicates that the method is best applied
to large documents.

In light of these considerations, the method described here may be
considered suplemental to more traditional author identification
methods. 

The method is naturally agnostic to the particular content of a
document --- it does not require selecting some subset of words to use
for comparisons --- and so should be applicable to documents across
different styles and genres.  The analysis could be applied to any
document amenable to statistical parsing.  (It does seem that
documents with a lot of specialized notation, such as mathematical or
chemical notation would require adaptation to the parser.)

This paper introduces many possibilities for future work.  Of course
there is the question of how this will apply to other work in author
identification.  It is curious that the dimension reduction behaves so
differently for the \emph{Federalist} and \emph{Sanditon} ---
\emph{Federalist } best in smaller dimensions, but \emph{Sanditon}
works better in larger dimensions.  Given recent furor over machine
learning, it would be interesting to see if the features extracted by
the grammatical parser correspond in any way to features that would be
extracted by a ML tool.  (My suspicion is that training on current ML
tools does not extract grammatical information applicable to the
author identification problem.)

\appendix

\section{A Brief Introduction to Statistical Parsing}

At the suggestion of an anonymous reviewer, this appendix was written
to provide a brief a discussion of the statistical parsing, drawing
very closely from \cite[Chapter 14, Statistical
Parsing]{JurafskyMartin2009}.  More detailed discussions are provided
in \cite{KleinManning2003,KleinManning2003PCFG}.  The probabilistic
grammar employed is a probabilistic context-free grammar (PCFG).  In
this grammar are rules for transforming nonterminal symbols to a
string of symbols which could be nonterminal symbols or terminal
symbols.  In a PCFG each rule is accompanied by a probability.  As an
example, Table \ref{tab:pcfg1} shows a grammar for a toy language
(used for airline reservations).  Each rule in the table 
of the form
$A \rightarrow B \quad  [p]$
means that $p$ is the probability that the non-terminal $A$ will be
expanded to the sequence $B$.  This can be alternatively represented
as $P(A \rightarrow B)$ or as $P(A \rightarrow B | A)$ or as
$P(\text{LHS} | \text{RHS})$, where LHS and RHS mean ``left hand
side'' and ``right hand side,'' respectively.

In Table \ref{tab:pcfg1}, S denotes an start symbol (for a sentence).
The grammar's first rule says that a sentence may consist of a NP
(noun phrase) followed by a VP (verb phrase), and that such a rule
occurs with probability 0.8.  The second rule says that a sentence may
consist of an auxiliary (such as \emph{does} or \emph{can}) followed
by NP then a VP, with probability 0.15.  The next rule says that a
sentence can be a verb phrase, VP.  The tokens obtained by application
of a rule can be recursively expanded using their own rules, shown in
the table.  Thus, a NP may consist of a pronoun, or a proper-noun,
etc.  The probabilities are estimated from a large corpus of data
parsed by a linguist.

There is also a lexicon (or dictionary) of terms, each term of which
has a probability within its lexicon type.  The right side of Table
\ref{tab:pcfg1} illustrates a lexicon for this application.  For
example, a determiner \emph{Det} can be \emph{that} (with probability
0.1) or \emph{a} (with probability 0.3) or \emph{the} (with
probability 0.6).  A noun (in this application context) can be book,
flight, meal, money, flights, or dinner, with their respective
probabilities.  Probabilities are (again) estimated from a corpus.

A PCFG can be used to estimate the probability of a parse tree, which
can be used to disambiguate multiple parsings, or it can be used to
determine the probability of a sentence in a language modeling setting.

\begin{table}
  \caption{Example rules for a PCFG (see Figure 14.1 of
    \cite{JurafskyMartin2009}).  S=start symbol (or sentence);
    NP=noun phrase; VP = verb phrase; PP=prepositional phrase.)
}
  \centering
  \begin{tabular}{|ll||l|}
    \hline
    
\textbf{Grammar} & \textbf{Probability} & \textbf{Lexicon} \\
\hline
S $\rightarrow$ NP~ VP     & [0.80] & \emph{Det} $\rightarrow$
                                   \textit{that} [0.10] $|$ \textit{a}
                                   [.30]  $|$ \textit{the} [.60] \\
S $\rightarrow$ \emph{Aux}~ NP~NP & [0.15] & \textit{Noun} $\rightarrow$
                                     \emph{book} [.10] $|$
                                             \emph{flight} [.30] $|$ \\
S $\rightarrow$ VP & [0.05] & \qquad \emph{meal} [.15] $|$ \emph{money} [.05]
    \\
NP $\rightarrow$ \emph{pronoun}  & [0.35]  & \qquad \emph{flights} [.40] $|$
                                        \emph{dinner} [.10] \\
    NP $\rightarrow$ \emph{Proper-Noun} & [0.30] & 
\emph{Verb} $\rightarrow$ \emph{book} [0.30] $|$ \emph{include} [0.30]
                                                   $|$ 
\\
    NP $\rightarrow$ \emph{Det}~\emph{Nominal} & [0.20]&  \qquad \emph{prefer} [0.40]
\\
    NP $\rightarrow$ \emph{Nominal} & [0.15] & 
\emph{Prounoun} $\rightarrow$ \emph{I} [0.40] $|$ \emph{you} [0.40] $|$
\\
    \emph{Nominal} $\rightarrow$ \emph{Noun} & [0.75] &
\qquad \emph{me} [0.15] $|$ \emph{you} [0.40]
\\
    \emph{Nominal} $\rightarrow$ \emph{Nominal}~\emph{Noun} & [0.20] &
\emph{Proper-noun} $\rightarrow$ \emph{Houston} [0.60] $|$
\\
    \emph{Nominal} $\rightarrow$ \emph{Nominal}~PP & [0.05] &
\qquad \emph{NWA} [0.40]
 \\
    VP $\rightarrow$ \emph{Verb} & [0.35] &
\emph{Aux} $\rightarrow$ \emph{does} [0.60] $|$ \emph{can} [0.40]
\\
VP $\rightarrow$ \emph{Verb}~NP & [0.20] & 
\emph{Preposition} $\rightarrow$ \emph{from} [0.30] $|$ \emph{to}
                                           [0.30] $|$
\\
    VP $\rightarrow$ \emph{Verb}~NP~PP & [0.10] & 
\qquad \emph{on} [0.20] $|$ \emph{near} [0.15] $|$ \\
    VP $\rightarrow$ \emph{Verb}~PP & [0.15] & 
\qquad \emph{through} [0.15]
\\
    VP $\rightarrow$ \emph{Verb}~NP~PP & [0.05] & \\
    VP $\rightarrow$ VP~PP & [0.15] & \\
    PP $\rightarrow$ \emph{Preposition}~ NP & [1.0]  & \\
    \hline
  \end{tabular}
  \label{tab:pcfg1}
\end{table}

``The probability of a particular parse tree $T$ is defined as the
product of the probabilities of all the rules used to expand each of
the $n$ non-terminal nodes in the parse tree $T$, where each rule $i$
can be expressed as LHS$_i \rightarrow$ RHS$_i$:
\[ P(T,S) = \prod_{i=1}^n P(\mathtt{RHS}_i | \mathtt{LHS}_i)
\]
The resulting probability $P(T,S)$ is \ldots the joint probability of
the parse and the sentence and the probability of the parse $P(T)$.''
\cite[p. 462]{JurafskyMartin2009} .  In computing the probability on a
tree, there is a factor for every rule, which corresponds to every
edge on the tree.

As an example, consider two different ways of parsing the sentence:
``Book the dinner flight.''   
This can be parsed (understood) either as
\begin{quote}
  \circnum{1} Book a flight that serves dinner
\end{quote}
or as
\begin{quote}
  \circnum{2} Book a flight for [on behalf of]  the dinner.
\end{quote}
The parse tree, rules, and corresponding probability for parsing \circnum{1} is shown here:

\begin{tabular}{ll}
\begin{minipage}[t]{.25\textwidth}
                   \vspace*{0in}
\begin{forest}
  [S[VP[VERB[\emph{Book}]][NP[Det[\emph{the}]][Nominal
[Nominal[Noun[\emph{dinner}]]][Noun[\emph{flight}]]
]]]]
\end{forest}
\end{minipage}
&
 \begin{minipage}[t]{.6\textwidth}
    \vspace*{0in}
\begin{tabular}{llll}
  \multicolumn{3}{c}{Rules} & Prob \\
  \hline
  S            &$\rightarrow$& VP                  & .05 \\
  VP         &$\rightarrow$& Verb~NP          & .20  \\
  NP         &$\rightarrow$&  Det Nominal   & .20 \\
  Nominal &$\rightarrow$& Nominal Noun & .20 \\
  Nominal &$\rightarrow$& Noun              & .75 \\
  Verb      &$\rightarrow$& \emph{book}   & .30 \\
  Det        &$\rightarrow$& \emph{the}      & .60 \\
  Noun     &$\rightarrow$& \emph{dinner} & .10 \\
  Noun     &$\rightarrow$& \emph{flights} & .40\\                                  
\end{tabular}
~ \\
~\\
{\scriptsize
  $P(T_1)  = (.05)(.2)(.2)(.2)(.75)(.3)(.6)(.1)(.4) = 2.2\times 10^{-6}$
}
\end{minipage}
\end{tabular}

The parse tree, rules, and corresponding probability for parsing \circnum{2} is shown here:

\begin{tabular}{ll}
 \begin{minipage}[t]{.25\textwidth}
    \vspace*{0in}
  \begin{forest}
[S[VP[Verb[\emph{Book}]][NP[Det[\emph{the}]][Nominal[Noun[\emph{dinner}]]]]
[NP[Nominal[Noun[\emph{flight}]]]]]]
\end{forest}
\end{minipage}
&  
\begin{minipage}[t]{.6\textwidth}
    \vspace*{0in}
\begin{tabular}{llll}
  \multicolumn{3}{c}{Rules} & Prob \\
  \hline
  S            &$\rightarrow$& VP                  & .05 \\
  VP  &$\rightarrow$& Verb~NP~NP  & .10 \\
  NP  &$\rightarrow$& Det~Nominal & .20 \\
  Nominal &$\rightarrow$& Noun & .75 \\
  Nominal &$\rightarrow$& Noun & .75 \\
  Verb      &$\rightarrow$& \emph{book}   & .30 \\
  Det        &$\rightarrow$& \emph{the}      & .60 \\
  Noun     &$\rightarrow$& \emph{dinner} & .10 \\
  Noun     &$\rightarrow$& \emph{flights} & .40  \\                                    
\end{tabular}
~ \\
~\\
{\scriptsize
$P(T_2)  = (.05)(.1)(.2)(.15)(.75)(.75)(.3)(.6)(.1)(.4) = 6.1\times 10^{-7}$
}
\end{minipage}
\end{tabular}

The probabilities computed for these two parse structures are
\[ P(T_1) = 2.2 \times 10^{-6} \qquad
  P(T_2) = 6.1 \times 10^{-7}.
\]
The parsing \circnum{1} has much higher probability than parsing \circnum{2} (which accords with a common understanding of the sense of the sentence). 

The parser works through the text being parsed, probabilistically
associating the word with its grammatical element, in the context of
the tree that is being built.  When competing trees are constructed,
the tree with highest probability is accepted.

\section{Dimension Reduction: Some Mathematical Details}

This material is drawn from \cite{HowlandJeonPark}.
The trace of $S_w$ provides a measure of the clustering of the feature for each class around their respective centroids,
\[ \trace(S_w) = \sum_{i=1}^k \sum_{j \in N_i} (\vbf_j - \cbf^{(i)})^T (\vbf_j - \cbf^{(i)}) =
  \sum_{i=1}^k \sum_{j \in N_i} \|   \vbf_j - \cbf^{(i)} \|^2.
\]
Note that $S_w$, being the sum of the outer product of $n$ terms,
generically has rank $\min(n,m)$.  In the work here, the dimension of
the feature vectors $m$ is very large, so that $\rank(S_w) = n$; $S_w$
is singular.

Similarly, $\trace(S_b)$ measures the total distance between cluster centroids and the overall centroid,
\[ \trace(S_b) = \sum_{i=1}^k \sum_{j \in N_i} (\cbf^{(i)} - \cbf)^T (\cbf^{(i)} - \cbf) = 
  \sum_{i=1}^k \sum_{j \in N_i} \|\cbf^{(i)} - \cbf)\|^2.
\]
A measure of cluster quality which measures the degree to which
$\trace(S_w)$ is small and $\trace(S_b)$ is large is
\[ J_1 = \trace(S_w^{-1} S_b)
\]
As noted above, $S_w$ is singular, so this a conceptual expression
(not actually computed).  In \cite{ZhangWuNiuDing2014}, the problem of
the singularity of $S_w$ is dealt with by working with a regularized
scatter matrix $S_w + \lambda I$, for some regularization parameter
$\lambda$, which is found there by searching over a range of
$\lambda$s which provide best performance.  The method described here
using the SVD avoids the need to perform this search (and the
possibility that some performance may have been sacrificed by not
finding an optimum value of $\lambda$).

When the vectors are transformed by the transformation $G^T$, the scatter matrices are
\[ S_{w,G} = \sum_{i=1}^k \sum_{j \in N_i} (G^T \vbf_j - G^T \cbf^{(i)}) (G^T \vbf_j - G^T \cbf^{(i)})^T = G^T S_w G,
\]
and (similarly) $S_{b,G} = G^T S_b G$ and $S_{m,G} = G^T S_m G$.
The goal now is to choose $G^T$ to make $\trace(S_{w,G})$ small while making $\trace(S_{b,G})$ large.  More precisely, the matrix $G$ is sought that maximizes
\[ J_1(G) = \trace((G^T S_w G)^{-1}(G^T S_b G)).
\]
In this case, the matrix $G^T S_w G$ may not be singular.

To express the algorithm, the following matrices are defined.  The
scatter matrices $S_w$, $S_b$ and $S_m$ can be expressed in terms of
the matrices
\[ H_w = \begin{bmatrix} 
    V_1 - \cbf^{(1)} \ebf_{n_1} &
    V_2 - \cbf^{(2)} \ebf_{n_2} & \cdots &
    V_k - \cbf^{(k)} \ebf_{n_k}
  \end{bmatrix} \in \Rbb^{\matsize{m}{n}},
\]
\[
  H_b = \begin{bmatrix}
    \sqrt{n_1}( \cbf^{(1)}  - \cbf) &
    \sqrt{n_2}( \cbf^{(2)}  - \cbf) & \cdots &
    \sqrt{n_k}( \cbf^{(k)}  - \cbf)    
 \end{bmatrix} \in \Rbb^{\matsize{m}{k}}
\]
and
\[ H_m = \begin{bmatrix}
    \vbf_1 - \cbf &
    \vbf_2 - \cbf &      \cdots &
    \vbf_n - \cbf 
 \end{bmatrix} \in \Rbb^{\matsize{m}{n}}
\]
as
\[ S_w = H_w H_w^T \qquad  S_b = H_b H_b^T \qquad
  S_m = H_m H_m^T.
\]
That is, $H_m$, $H_b$ and $H_m$ form factors of the respective scatter matrices.

The algorithm for computing $G$ is shown below.
(adapted from Algorithm 1 of \cite{HowlandJeonPark}).
\medskip

  \hrule \smallskip
\textbf{Algorithm 1} Finding a structure-preserving, dimension-reducing matrix $G$:
  \hrule \smallskip
\noindent Given matrices $H_b \in \Rbb^{\matsize{m}{k}}$ and $H_w \in \Rbb^{\matsize{m}{n}}$ (the factors of $S_b$ and $S_w$), determines the matrix $G \in \Rbb^{\matsize{m}{\ell}}$ which preserves the cluster structure in the $\ell$ dimensional space.

\noindent\textbf{Input:} $H_b$, $H_w$, $\ell$. \qquad \textbf{Output:} $G$

\begin{enumerate} \setlength{\itemsep}{0pt}
\item Form 
\[ K = \begin{bmatrix} H_b^T \\[5pt] H_w^T \end{bmatrix} \in \Rbb^{\matsize{(k+n)}{m}}
\]    
and compute its SVD
\[ K = P \begin{bmatrix} R & \zerobf \\ \zerobf & \zerobf \end{bmatrix} Q^T.
\]
Determine the rank of $K$:
\[ t = \rank(R)
\]
\item Compute the SVD of a submatrix of $P$:
  \[ P(1:m, 1:t) = U \Sigma W^T
  \]

\item   Form an empty matrix $G \in \Rbb^{\matsize{m}{\ell}}$.
\item (Compute $G$ as the first $\ell$ columns of $Q \begin{bmatrix} R^{-1} W & \zerobf \\ \zerobf & I \end{bmatrix}$.  This can be done as follows:) \\
  
(Overwrite the first $t$ columns of $Q$ as $QR^{-1}$:) \\
  for $j=1:t$ \\
\q   $Q(:,j) = Q(:,j)/R(j,j)$ \\
  end \\
  if($\ell \leq t$) \\
\q $G(:,1:\ell) = Q(:,1:t)W(:,1:\ell)$ \\
  else  \\
\q\q Print: ``Number of columns of $G$ requested exceeds number of
nontrivial \\
\q\q\q  singular values pairs of $H_b^T$ and $H_w^T$'' \\
\q\q $G(:,1:t) = Q(:,1:t)W(:,1:t)$ \\
\q if($\ell > n$) \\
\q\q Print: ``And it exceeds the number of columns of $G$'' \\
\q else  (Set the remaining columns of $G$ equal to $Q_2$) \\
\q\q $G(:,t+1:\ell) = Q(:,t+1,\ell)$ \\
\q end \\
end \\
\end{enumerate}
\hrule

\bibliographystyle{ieeetr}
\newcommand{\bibdir}{.}
\bibliography{\bibdir/bibliog,\bibdir/miscbib,\bibdir/textanal}

\end{document}